\PassOptionsToPackage{unicode}{hyperref}
\PassOptionsToPackage{hyphens}{url}
\PassOptionsToPackage{dvipsnames,svgnames,x11names}{xcolor}
\documentclass[
]{article}
\usepackage{xcolor}
\usepackage{amsmath,amssymb}
\usepackage{iftex}
\ifPDFTeX
  \usepackage[T1]{fontenc}
  \usepackage[utf8]{inputenc}
  \usepackage{textcomp} % provide euro and other symbols
\else % if luatex or xetex
  \usepackage{unicode-math} % this also loads fontspec
  \defaultfontfeatures{Scale=MatchLowercase}
  \defaultfontfeatures[\rmfamily]{Ligatures=TeX,Scale=1}
\fi
\usepackage{lmodern}
\ifPDFTeX\else
\fi
\IfFileExists{upquote.sty}{\usepackage{upquote}}{}
\IfFileExists{microtype.sty}{% use microtype if available
  \usepackage[]{microtype}
  \UseMicrotypeSet[protrusion]{basicmath} % disable protrusion for tt fonts
}{}
\makeatletter
\@ifundefined{KOMAClassName}{% if non-KOMA class
  \IfFileExists{parskip.sty}{%
    \usepackage{parskip}
  }{% else
    \setlength{\parindent}{0pt}
    \setlength{\parskip}{6pt plus 2pt minus 1pt}}
}{% if KOMA class
  \KOMAoptions{parskip=half}}
\makeatother
\makeatletter
\ifx\paragraph\undefined\else
  \let\oldparagraph\paragraph
  \renewcommand{\paragraph}{
    \@ifstar
      \xxxParagraphStar
      \xxxParagraphNoStar
  }
  \newcommand{\xxxParagraphStar}[1]{\oldparagraph*{#1}\mbox{}}
  \newcommand{\xxxParagraphNoStar}[1]{\oldparagraph{#1}\mbox{}}
\fi
\ifx\subparagraph\undefined\else
  \let\oldsubparagraph\subparagraph
  \renewcommand{\subparagraph}{
    \@ifstar
      \xxxSubParagraphStar
      \xxxSubParagraphNoStar
  }
  \newcommand{\xxxSubParagraphStar}[1]{\oldsubparagraph*{#1}\mbox{}}
  \newcommand{\xxxSubParagraphNoStar}[1]{\oldsubparagraph{#1}\mbox{}}
\fi
\makeatother

\usepackage{longtable,booktabs,array}
\usepackage{calc} % for calculating minipage widths
\usepackage{etoolbox}
\makeatletter
\patchcmd\longtable{\par}{\if@noskipsec\mbox{}\fi\par}{}{}
\makeatother
\IfFileExists{footnotehyper.sty}{\usepackage{footnotehyper}}{\usepackage{footnote}}
\makesavenoteenv{longtable}
\usepackage{graphicx}
\makeatletter
\newsavebox\pandoc@box
\newcommand*\pandocbounded[1]{% scales image to fit in text height/width
  \sbox\pandoc@box{#1}%
  \Gscale@div\@tempa{\textheight}{\dimexpr\ht\pandoc@box+\dp\pandoc@box\relax}%
  \Gscale@div\@tempb{\linewidth}{\wd\pandoc@box}%
  \ifdim\@tempb\p@<\@tempa\p@\let\@tempa\@tempb\fi% select the smaller of both
  \ifdim\@tempa\p@<\p@\scalebox{\@tempa}{\usebox\pandoc@box}%
  \else\usebox{\pandoc@box}%
  \fi%
}
\def\fps@figure{htbp}
\makeatother

\usepackage[]{natbib}
\PassOptionsToPackage{numbers}{natbib}
\usepackage[final,dblblindworkshop]{neurips_2026}
\usepackage{placeins}
\workshoptitle{2nd Workshop on Advances in Representation Learning for Earth Observation}

\AtBeginDocument{\author{Johannes Hirn\\
{\small Image Processing Laboratory (IPL), Universitat de València, 46980 Paterna, Valencia, Spain}\\
{\small Universidad Cardenal Herrera-CEU, 46115 Alfara del Patriarca, Valencia, Spain}}}
\makeatletter
\@ifpackageloaded{caption}{}{\usepackage{caption}}
\AtBeginDocument{%
\ifdefined\contentsname
  \renewcommand*\contentsname{Table of contents}
\else
  \newcommand\contentsname{Table of contents}
\fi
\ifdefined\listfigurename
  \renewcommand*\listfigurename{List of Figures}
\else
  \newcommand\listfigurename{List of Figures}
\fi
\ifdefined\listtablename
  \renewcommand*\listtablename{List of Tables}
\else
  \newcommand\listtablename{List of Tables}
\fi
\ifdefined\figurename
  \renewcommand*\figurename{Figure}
\else
  \newcommand\figurename{Figure}
\fi
\ifdefined\tablename
  \renewcommand*\tablename{Table}
\else
  \newcommand\tablename{Table}
\fi
}
\@ifpackageloaded{float}{}{\usepackage{float}}
\floatstyle{ruled}
\@ifundefined{c@chapter}{\newfloat{codelisting}{h}{lop}}{\newfloat{codelisting}{h}{lop}[chapter]}
\floatname{codelisting}{Listing}

\makeatother
\makeatletter
\@ifpackageloaded{caption}{}{\usepackage{caption}}
\@ifpackageloaded{subcaption}{}{\usepackage{subcaption}}
\makeatother
\usepackage{bookmark}
\IfFileExists{xurl.sty}{\usepackage{xurl}}{} % add URL line breaks if available
\makeatletter
\@ifundefined{xmpquote}{\newcommand{\xmpquote}[1]{#1}}{}
\makeatother
\hypersetup{
  pdftitle={Emergent Sparsity: How Many Effective Variables Survive Coarse-Graining in AlphaEarth Embeddings?},
  pdfauthor={Johannes Hirn},
  pdfkeywords={\xmpquote{AlphaEarth}, \xmpquote{coarse-graining}, \xmpquote{emergence}, \xmpquote{effective
dimension}, \xmpquote{rate-distortion}, \xmpquote{beta-VAE}, \xmpquote{spatial
aggregation}, \xmpquote{Earth observation}},
  colorlinks=true,
  linkcolor={blue},
  filecolor={Maroon},
  citecolor={Blue},
  urlcolor={Blue},
  pdfcreator={LaTeX via pandoc}}

\title{Emergent Sparsity: How Many Effective Variables Survive
Coarse-Graining in AlphaEarth Embeddings?}
\author{Johannes Hirn}
\date{}
\begin{document}
\maketitle
\begin{abstract}
AlphaEarth encodes each 10 m land pixel as a 64-component vector
intended to support diverse global mapping tasks. Yet many scientific
applications work at coarser spatial supports, for which fewer than 64
effective variables may suffice. We ask whether these vectors can be
reorganized into ranked global nonlinear coordinates, how many suffice
at a chosen reconstruction accuracy, and whether that number decreases
with spatial scale. We thus describe effective dimension as a
two-control surface over tolerance and spatial scale, rather than one
intrinsic number. At each scale, a \(\beta\)-VAE scan with a common
architecture produces a rank--distortion envelope, giving the smallest
retained count achieved within the scanned family at each reconstruction
tolerance. At 3\% validation tolerance, this number falls from 24 to 7
when the scale is increased from 10 m to 78.9 km, whereas the
corresponding PCA rank falls only from 46 to 38. Repeating the analysis
on unit-norm targets preserves this trend.
\end{abstract}

\subsection{Introduction}\label{introduction}

AlphaEarth summarizes diverse Earth observations in a 64-component
vector at each 10 m land pixel \citep{brown2025alphaearth}. Many
applications instead work at scales from hundreds of meters to tens of
kilometers, where fewer effective variables may suffice.

The idea that fewer variables suffice at coarser resolution has a long
history in physics, where effective descriptions do not track
microscopic degrees of freedom individually, but retain their collective
influence in macroscopic variables or effective
couplings---thermodynamics is the familiar example. At any fixed
resolution, however, an effective theory is also specified by its
required accuracy: a cutoff fixes the shortest resolved length scale,
while the desired accuracy determines how far the effective expansion
must be carried \citep{weinberg1979phenomenological}.

In our setting, the corresponding controls are spatial support and
tolerated reconstruction distortion. Because greater sparsity generally
comes at a cost in accuracy, an effective-variable count is meaningful
only once we specify both scale and reconstruction tolerance separately.
Spatial scale \(\ell\) is the side length of a Euclidean square with the
same area as the averaging region. \(\ell\) changes the target before
any model fitting, while \(\varepsilon\) sets the acceptable
reconstruction distortion.

The tolerance dependence can already be examined at the native 10 m
scale, before any coarse-graining. AlphaEarth uses a fixed-concentration
von Mises--Fisher bottleneck, with each vector specifying a mean
direction on \(S^{63}\subset\mathbb{R}^{64}\). A batch-uniformity
objective promotes broad sphere utilization to avoid representation
collapse---vectors clustering in a restricted region---and improve
downstream performance \citep{brown2025alphaearth}. Yet broad sphere
utilization does not imply that 63 degrees of freedom are required for
finite-accuracy reconstruction, so we test whether \(\beta\)-VAEs can
reconstruct native embeddings at a given tolerance using fewer retained
coordinates.

In contrast to AlphaEarth's vMF bottleneck, with a single fixed,
nonlearnable concentration for its 64-component directional
representation, \(\beta\)-VAEs learn diagonal-Gaussian posteriors with
input-dependent variances for each latent coordinate. The regularization
strength \(\beta\) controls the trade-off between reconstruction and KL
cost \citep{kingma2014autoencoding, higgins2017betavae}, acting as an
effective temperature \citep{hirn2026posteriorcollapse}: increasing it
favors coordinate-wise posterior collapse as retaining information
becomes more costly \citep{hirn2026generative}. Validation KL rate
orders the coordinates by information cost, while truncation tests how
accurately successive retained sets reconstruct. Scanning \(\beta\) thus
produces a rank--distortion envelope at each scale.

We then turn to the scale dependence of this tolerance-dependent count.
Here again, physics provides the inspiration: the renormalization group
describes how effective descriptions change as resolution is reduced,
distinguishing perturbations that grow from those that fade
\citep{wilson1971rg1}. Our analogy is narrower: because each model is
trained independently, we ask only whether fewer variables suffice at
larger scales, not whether the same variables persist along a
renormalization-group flow.

Spatial averaging also provides a geometric control: native embeddings
lie on the unit sphere, whereas their means generally lie inside the
corresponding ball---coarse-graining therefore changes both direction
and radius. We perform scans of the KL weight \(\beta\) before and after
rescaling each mean to unit norm, testing whether dimensional reduction
persists when mean-vector norms are fixed to one.

Existing studies provide complementary views of AlphaEarth's
dimensionality. In an analysis of the contiguous United States at native
resolution, \citet{rahman2026characterizing} report a participation
ratio of 13.3 and a local intrinsic dimension near 10. Extending this
scalar view across scales, \citet{rao2026intrinsicdimensionearth} find a
slight decrease in four architecture-agnostic estimates under circular
averaging from 20 to 500 m. \citet{rahman2026alphaearthinterpretable}
and \citet{benavides2026whatonearthisalphaearth} interpret AlphaEarth
through environmental probes or particular downstream tasks at specific
scales. Instead we learn global nonlinear coordinates without semantic
labels and measure how many suffice across scales and tolerances.

\subsection{Data and method}\label{data-and-method}

\subsubsection{Matched scales}\label{matched-scales}

We analyze 21,299 matched land anchors across scales from 10 m to 78.9
km. The anchors lie at the centers of level-6 icosahedral-mesh
triangles: we retain sites whose land-mask proxy classifies at least 61
of their 64 level-9 descendant centers (95.3\%) as land. Every scale
uses the same anchor IDs and geographically blocked seed-0 split: each
whole level-4 parent triangle, roughly 441 km on a side, belongs to one
subset, so nearby anchors within the same parent cannot cross
train/validation/test boundaries. This yields 14,881 training, 3,154
validation, and 3,264 test anchors.

For each anchor we retain its native AlphaEarth embedding from the year
2024 and form equal-weight means at several nested scales. For each
mean, we request 256 descendant-triangle centers four mesh levels below
its averaging polygon. Missing values are omitted, and repeated hits on
the same native raster pixel are counted once. At supports of 308 m and
above, more than 99.7\% of anchors contribute all 256 samples. These
averages approximate the mean over all native pixels in each polygon.

\subsubsection{Effective variables and
dimension}\label{effective-variables-and-dimension}

At each scale we scan a \(\beta\)-VAE with 128 latent coordinates. At a
given \(\beta\), some coordinate posteriors approach the prior and carry
little input information (posterior collapse), while others remain
informative. We rank all 128 coordinates by decreasing validation KL
rate, \(R_j=\mathbb{E}_x[D_{\mathrm{KL}}(q(z_j\mid x)\|p(z_j))]\), which
measures information cost rather than direct reconstruction utility. For
rank \(K\), we decode the first \(K\) posterior means and set the
remaining coordinates to zero, the prior mean. Each retained
posterior-mean function \(\mu_j(x)\) is a candidate effective variable:
a learned global coordinate whose reconstruction utility, including
interactions, is tested rather than inferred from its rate alone.

For each retained count \(K\), we take the lowest squared-error
validation distortion across the scanned \(\beta\) values. Different
ranks may select different \(\beta\) values, so this envelope does not
represent nested truncations of a single fitted basis. The contours in
Figure~\ref{fig-dimension-support} report the lowest distortion achieved
with at most \(K\) coordinates.

At each scale and target geometry, the reference predictor is the mean
of all training targets, fixed before evaluation. Relative distortion is
\(D/D_0\), with \(D\) the mean squared Euclidean reconstruction error
and \(D_0\) the reference predictor's mean squared error on the same
evaluation observations. VAE and PCA use this reference rule.

We estimate effective dimension as the smallest retained count meeting
validation tolerance \(\varepsilon\). This is the smallest rank achieved
by the scanned pipeline, hence an empirical upper bound on the unknown
optimal dimension for these sampled targets and this distortion metric,
not proof of a global minimum. The 256-point approximation may retain
fluctuations that denser polygon sampling would suppress, while the
model family and optimization may prevent us from reaching the minimum.
Finite validation samples add uncertainty, and some validation-selected
models exceed the requested tolerance on test.

Input-space MSE retains polygon-mean radius as part of the
coarse-grained target. As a controlled sensitivity check, we repeat each
scan on targets rescaled to unit norm, keeping the architecture, loss,
data split, and \(\beta\) grid fixed. Means of unit vectors lie in the
closed unit ball, with radius recording directional coherence: the
unit-norm control removes this variation while retaining the mean
direction.

At every scale, both target geometries use a standard Gaussian latent
prior and the same fully connected \(\beta\)-VAE architecture, with
eight 256-unit hidden layers in each encoder and decoder, 128 latent
coordinates, and 26 independently optimized normalized \(\beta\) values
from \(10^{-6}\) to \(10^{-1}\). The envelope, rate order, and tolerance
rank are selected on validation and frozen before held-out test
evaluation. PCA follows the same train/validation/test protocol.

\subsection{Results}\label{results}

\begin{figure}[htbp]

\centering{

\includegraphics[width=1\linewidth,height=\textheight,keepaspectratio]{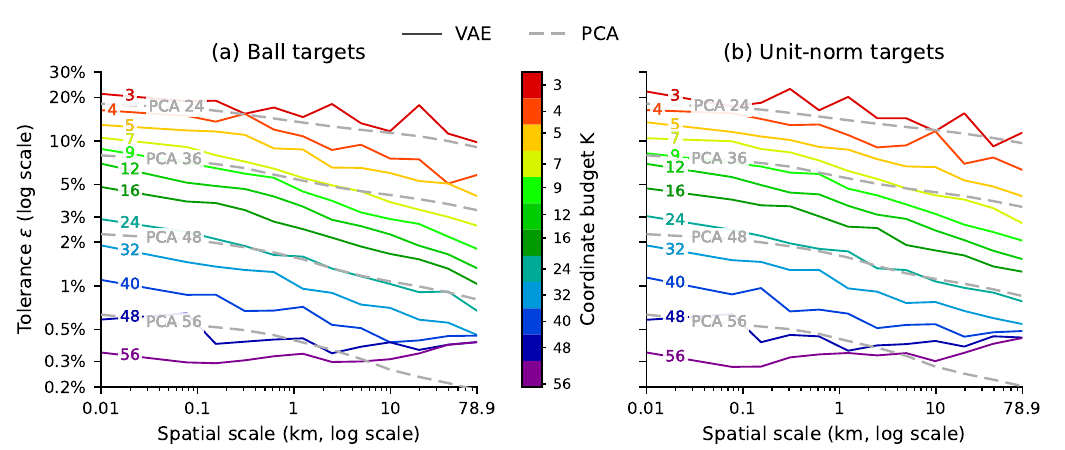}

}

\caption{\label{fig-dimension-support}Validation scale--tolerance
contours for ball (a) and unit-norm (b) targets. Each labeled \(K\)
curve gives the lowest relative distortion attained with at most \(K\)
coordinates. Colored solid curves are VAE envelopes over the \(\beta\)
scan, and gray dashed curves show PCA. Distortions are normalized by the
training-mean predictor's validation distortion at each support.}

\end{figure}

The number of selected effective variables decreases overall with scale
(Figure~\ref{fig-dimension-support}). From 10 m to 78.9 km, the VAE
count selected at 10\% validation tolerance falls from 8 to 3, at 3\%
from 24 to 7, and at 1\% from 41 to 18. At the same tolerances, the
corresponding PCA ranks fall from 33 to 23, 46 to 38, and 54 to 47.
Because relative distortion uses a scale-specific denominator, these
comparisons measure relative compressibility rather than the raw
attenuation caused by averaging. The contours show that scale does not
determine effective dimension by itself. Stringent reconstruction
remains high-dimensional, and for \(K=56\), PCA overtakes VAE at coarse
supports.

For the dimension comparison, we hold tolerance fixed and read the
corresponding rank at each support
(Figure~\ref{fig-appendix-tolerance-slices}). At 3\%, coarse-graining
reduces the VAE count by about 71\%, versus 17\% for PCA. The VAE
\(K=24\) contour approximately tracks PCA \(K=48\): comparable accuracy
is attained with half as many nonlinear coordinates.

Simplification persists after removing variation in vector norm. At 78.9
km, the ball and unit-norm controls both select 7 latents at a 3\%
validation tolerance. At 1\% they select 18 and 19, respectively. Our
main results are thus preserved when removing target-radius variation.

\subsection{Discussion}\label{discussion}

Instead of a single intrinsic dimension, our central result is a surface
\(d_{\mathrm{eff}}(\varepsilon,\ell)\): spatial scale changes the object
being represented through averaging, whereas tolerance changes how
faithfully that object must be reconstructed. These controls have
distinct physical meanings, even where their effects admit an
approximate scaling relation. This framework lets us ask which
reconstruction accuracy can be maintained as fewer variables describe
progressively larger regions.

At any chosen \((\varepsilon,\ell)\), the selected checkpoint provides
more than a dimension count: its encoder defines nonlinear coordinates
and its decoder reconstructs embeddings. The per-coordinate KL rates
provide an information-cost ranking, while the rank--distortion curve
for that checkpoint measures the reconstruction utility of successive
retained sets.

For the three-coordinate model at 78.9 km and 10\% tolerance,
exploratory environmental probes associate the first coordinate with
temperature and evaporative stress, and the third with precipitation,
moisture, evaporation, and carbon fluxes
(Figure~\ref{fig-appendix-k3-maps} and Figures
\ref{fig-appendix-k3-climate}--\ref{fig-appendix-k3-biosphere}). These
overlapping patterns suggest tentative thermal--aridity and
water--productivity interpretations, respectively. The second coordinate
is less clear. In a similar analysis of MusicVAE's latent space, leading
coordinates each summarized several human-defined rhythm or pitch
features, while later ranks had less clear individual meanings
\citep{barenboim2024music}. Studying what each later coordinate adds in
combination with earlier ones may clarify its meaning.

The unit-norm results show that directional compressibility also changes
with scale. The VAE--PCA gap in this control shows that nonlinear
coordinates describe this directional structure with fewer variables at
moderate reconstruction tolerances.

The validation contours in Figure~\ref{fig-dimension-support} suggest
that relative distortion approximately separates into a function of
retained count \(K\) multiplied by a spatial factor \(g(\ell)\). For
tolerance crossings within this scaling regime, the effective dimension
therefore depends approximately on a single rescaled tolerance:
\begin{equation}\protect\phantomsection\label{eq-effective-dimension-scaling}{
d_{\mathrm{eff}}(\varepsilon,\ell)\approx
F\!\left(\frac{\varepsilon}{g(\ell)}\right).
}\end{equation} Dividing the rank--distortion curves by the measured
spatial factor brings them onto a common profile for \(K=5\)--30,
producing approximate data collapse across supports
(Figure~\ref{fig-appendix-curve-collapse} and
Figure~\ref{fig-appendix-rank-collapse}). Between 100 m and 10 km,
\(g(\ell)\propto(\ell/\ell_0)^{-\gamma}\) approximately, with reference
scale \(\ell_0=1\,\mathrm{km}\).

Our experiments cover only an intermediate scale range. Below
AlphaEarth's native 10 m support lies an unresolved short-distance (UV)
hierarchy---from organisms to molecules---that can influence the
observations encoded by AlphaEarth. At the opposite long-distance (IR)
end, spatial averaging approaches a whole-globe description. An
interesting question for future work is to find the scales at which
variables can be dispensed with, and whether those crossovers match
characteristic physical scales.

Because each model is trained independently, the present analysis counts
effective variables but does not identify a renormalization group-like
flow of named variables---this would require cross-scale alignment.
After such alignment, persistence or loss of reconstruction utility
could provide an empirical analogue of relevance or irrelevance.
Interpreting such a flow physically would require environmental probes
across scales, beyond those at the coarsest support examined here.

Our reported dimension count remains an empirical, model-dependent
reconstruction requirement. The multiscale analysis uses one
architecture and seed, and tolerance crossings may be sensitive to
optimization variability, particularly on shallow curves at strict
tolerances. At the coarse supports, approximately the same number of
samples is averaged over progressively wider neighborhoods, so the scale
trend probes how their joint distribution changes with spatial
separation. The environmental probes interpret the coarsest support only
and do not establish stable semantic identities across scales.

\textbf{Conclusion.} Under coarse-graining of AlphaEarth embeddings, a
sparse nonlinear description suffices at moderate reconstruction
tolerances, with a larger proportional reduction in required count than
PCA. Stringent reconstruction retains a high-dimensional tail. At each
chosen scale and tolerance, the selected model provides candidate
nonlinear coordinates, the corresponding fitted encoder and decoder, and
a coordinate ranking. Exploratory coarse-scale associations suggest
broad environmental interpretations, but do not establish unique
physical meanings or semantic identities across scales. There is no
scale-only effective dimension: what survives depends on the required
accuracy.

\clearpage

\section*{References}\label{references}
\addcontentsline{toc}{section}{References}

\renewcommand{\bibsection}{}

\clearpage
\appendix
\setcounter{figure}{0}
\renewcommand{\thefigure}{A\arabic{figure}}
\setcounter{table}{0}
\renewcommand{\thetable}{A\arabic{table}}
\setcounter{equation}{0}
\renewcommand{\theequation}{A\arabic{equation}}
\renewcommand{\theHequation}{A.\arabic{equation}}

\section{Supplementary diagnostics}\label{supplementary-diagnostics}

\subsection{Input geometry across
scales}\label{input-geometry-across-scales}

\begin{figure}[htbp]

\centering{

\includegraphics[width=0.72\linewidth,height=\textheight,keepaspectratio]{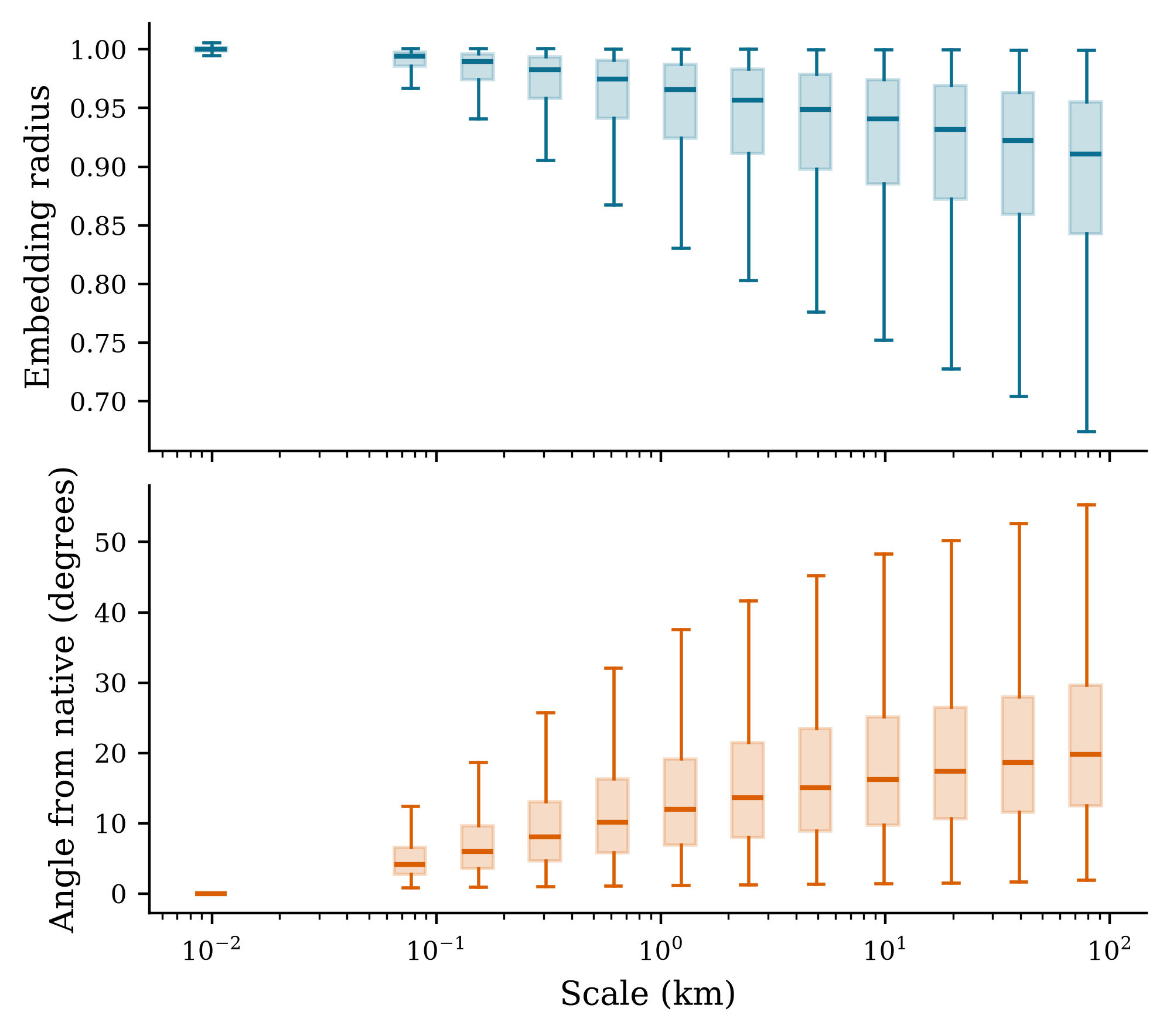}

}

\caption{\label{fig-appendix-geometry}Embedding radius and angle from
the matched native vector across scales. Boxes span the interquartile
range with a line at the median. Whiskers extend to the most extreme
observations within 1.5 interquartile ranges of the box, and individual
outliers are omitted.}

\end{figure}

Spatial averaging changes the input geometry before any VAE is fitted
(Figure~\ref{fig-appendix-geometry}). From 10 m to 78.9 km, median
radius decreases from 1.000 to 0.911, while the median angular
displacement from the matched 10 m vector grows from approximately zero
to \(19.8^\circ\). Coarse-graining therefore changes both the magnitude
and direction of the embedding field.

The boxes summarize the 14,881 training anchors: outliers are included
when computing the distributions but are not drawn individually.

\FloatBarrier

\clearpage

\subsection{Native-scale architecture
control}\label{native-scale-architecture-control}

\begin{figure}[htbp]

\centering{

\includegraphics[width=0.75\linewidth,height=\textheight,keepaspectratio]{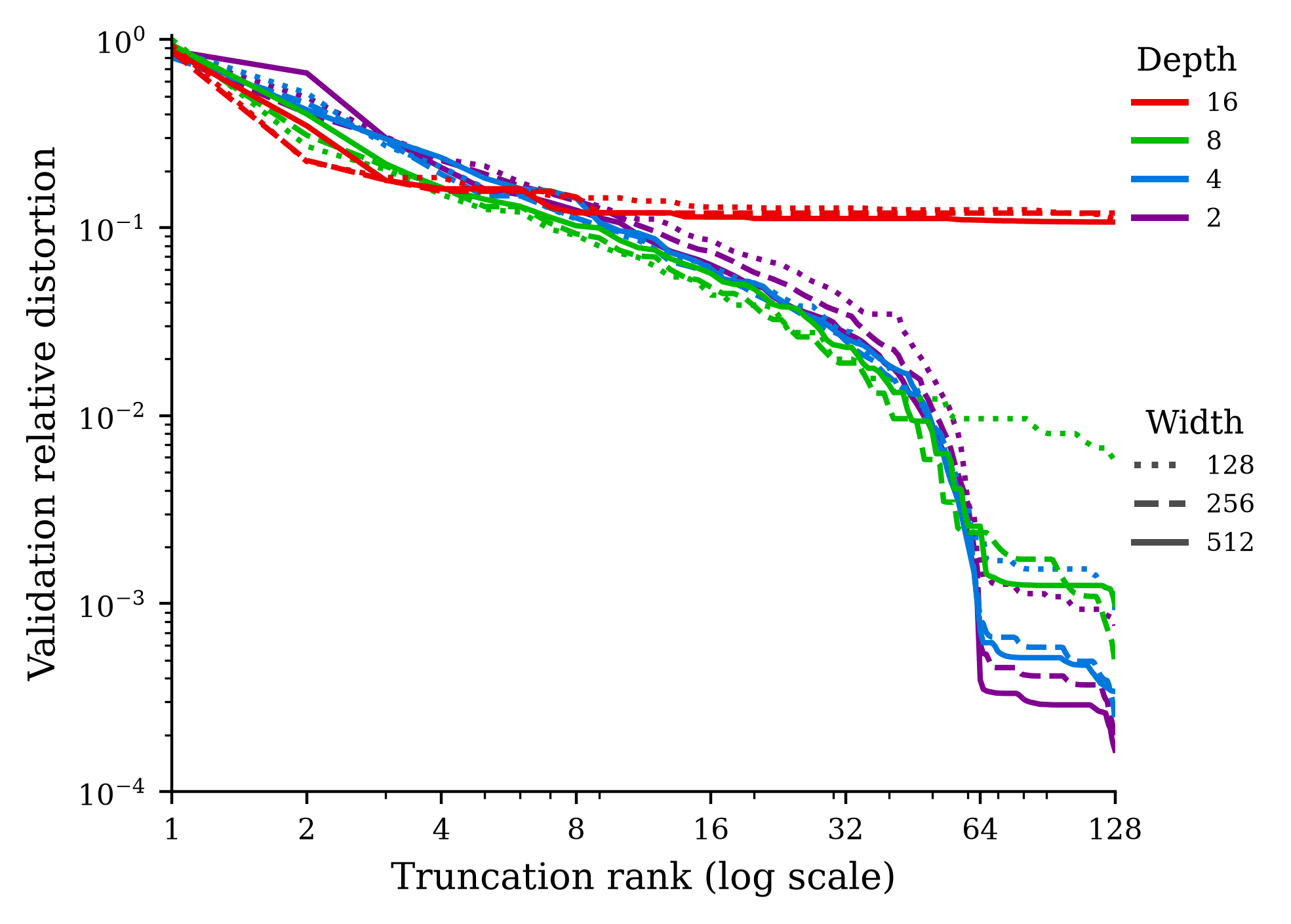}

}

\caption{\label{fig-appendix-architecture}Validation rank--distortion
curves at native 10 m support for the geography-aware split. Color
denotes encoder/decoder depth and line style denotes hidden width. Every
model has 128 latent coordinates.}

\end{figure}

We compare networks on validation in
Figure~\ref{fig-appendix-architecture} because architecture is a
model-selection choice. We select \texttt{8x256+128} to resolve the
stricter-tolerance tail: it requires the fewest coordinates among these
runs at 1\% and 0.3\%, with ranks 41 and 57, and ties with
\texttt{8x128+128} at rank 24 for 3\%. It is not best at every
tolerance: \texttt{8x128+128} reaches 10\% with seven coordinates,
compared with eight for the selected network.

\FloatBarrier

\clearpage

\subsection{\texorpdfstring{Two \(\beta\) scans, rank by
rank}{Two \textbackslash beta scans, rank by rank}}\label{two-beta-scans-rank-by-rank}

All reported multiscale curves use the same fully connected
\(\beta\)-VAE architecture, denoted \texttt{8x256+128}: the encoder and
decoder each have eight hidden layers of 256 units, and the stochastic
bottleneck has 128 coordinates. We use SiLU activations, batch size 256,
seed 0, and 26 independently optimized normalized \(\beta\) values from
\(10^{-6}\) to \(10^{-1}\). The best checkpoint for each retained count,
its latent order, and the count meeting each tolerance are chosen on
validation and frozen before test evaluation. Reconstruction is then
evaluated on the 3,264 held-out test anchors. PCA analogously uses train
for fitting, validation for rank selection, and test for evaluation.

Every VAE run uses Adam with learning rate \(3\times10^{-4}\) for at
most 50,000 parameter updates, stopping after 2,500 updates without
improvement in the validation objective. We restore the checkpoint with
the lowest validation reconstruction-plus-KL objective for that
\(\beta\). The subsequent rank-dependent selection compares these
restored checkpoints using validation reconstruction distortion.

\begin{figure}[htbp]

\centering{

\includegraphics[width=0.96\linewidth,height=\textheight,keepaspectratio]{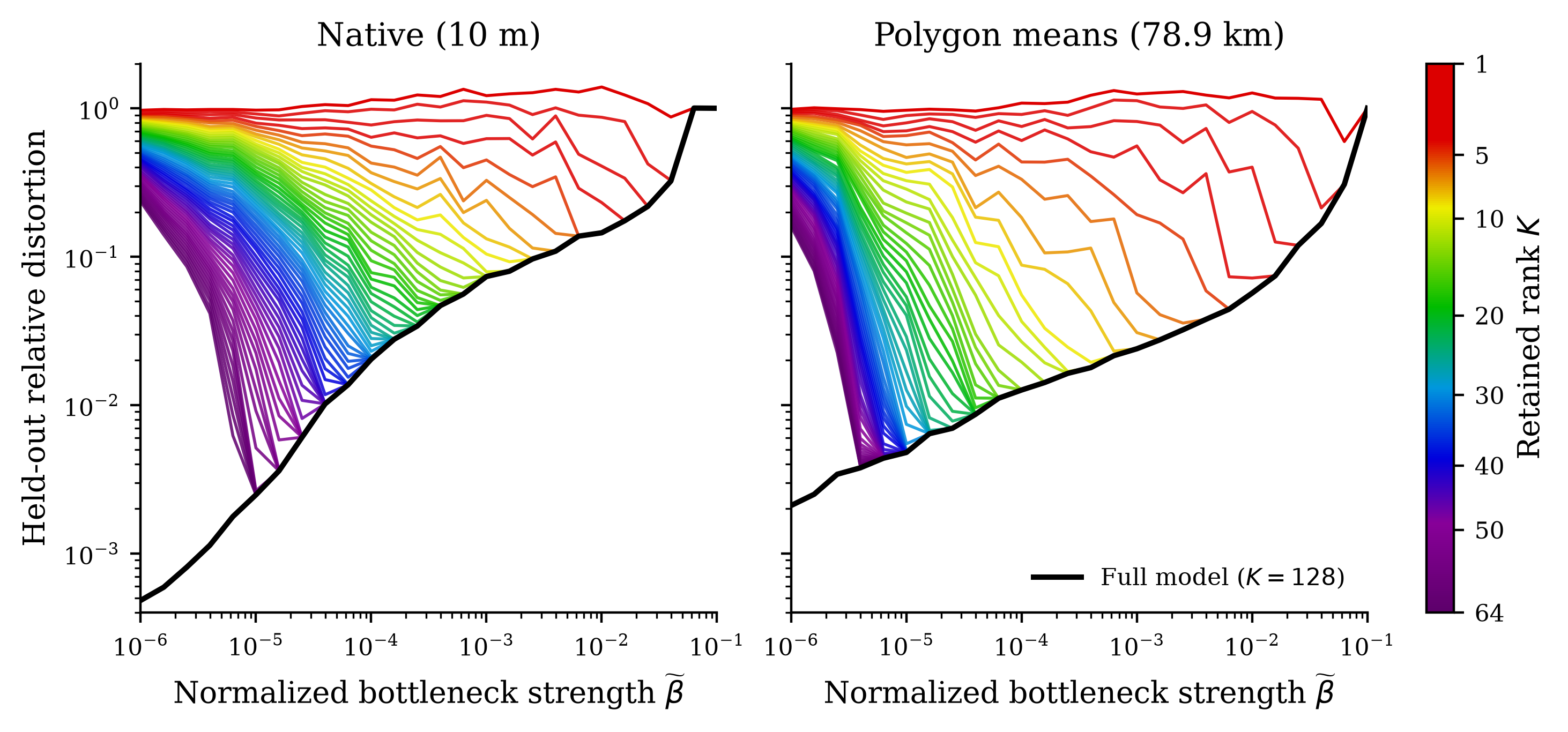}

}

\caption{\label{fig-appendix-beta}Held-out relative distortion for ball
targets across normalized \(\widetilde{\beta}\) at native 10 m support
(left) and for 78.9 km polygon means (right). Color denotes retained
ranks \(K=1\)--63, from red to dark violet. Black shows the full model
with all 128 coordinates retained across the entire scan.}

\end{figure}

In Figure~\ref{fig-appendix-beta}, coordinates are ordered by their
validation KL rates at each \(\beta\), and each curve fixes the retained
rank \(K\) while \(\beta\) varies. The panels report the resulting
relative distortions on the held-out test anchors at native 10 m support
and after averaging over 78.9 km polygons. Within each panel, all
checkpoints and retained ranks use the same training-mean reference
distortion on test, independent of \(\beta\) and \(K\). They expose
selected ranked truncation curves behind the rank--distortion envelopes:
test results are used only for reporting, never to select \(\beta\), the
latent order, or \(K\). A large improvement when \(K\) increases
identifies a rank and regularization regime in which an additional
variable has high reconstruction utility.

The decoder predicts the observation mean under a fixed isotropic
Gaussian observation model with variance \(1\) in every embedding
coordinate. Up to the additive Gaussian constant, the fitted objective
and normalized control parameter are therefore

\begin{equation}\protect\phantomsection\label{eq-vae-objective}{
\mathcal{L}_\beta
=\mathbb{E}\!\left[\frac{1}{2}\lVert x-\hat{x}\rVert^2\right]
+\beta\sum_j R_j,
\qquad
\widetilde{\beta}=\frac{\beta}{V_{\mathrm{train}}},
\qquad
V_{\mathrm{train}}
=\mathbb{E}_{\mathrm{train}}\!\left[\lVert x-\bar{x}_{\mathrm{train}}\rVert^2\right],
}\end{equation}

where \(R_j=\mathbb{E}_x[D_{\mathrm{KL}}(q(z_j\mid x)\|p(z_j))]\). Thus
\(V_{\mathrm{train}}\) is the training-mean squared error per sample,
computed separately for each scale and target geometry. The scan spans
\(\widetilde{\beta}=10^{-6},\ldots,10^{-1}\). At each scale, the
coefficient used in the training loss is
\(\beta=\widetilde{\beta}V_{\mathrm{train}}\). Plotted relative
distortions use the same fixed predictor's distortion on the evaluation
split as their denominator.

\FloatBarrier

\clearpage

\subsection{Held-out evaluation of the validation-selected
envelopes}\label{held-out-evaluation-of-the-validation-selected-envelopes}

\begin{figure}[htbp]

\centering{

\includegraphics[width=0.72\linewidth,height=\textheight,keepaspectratio]{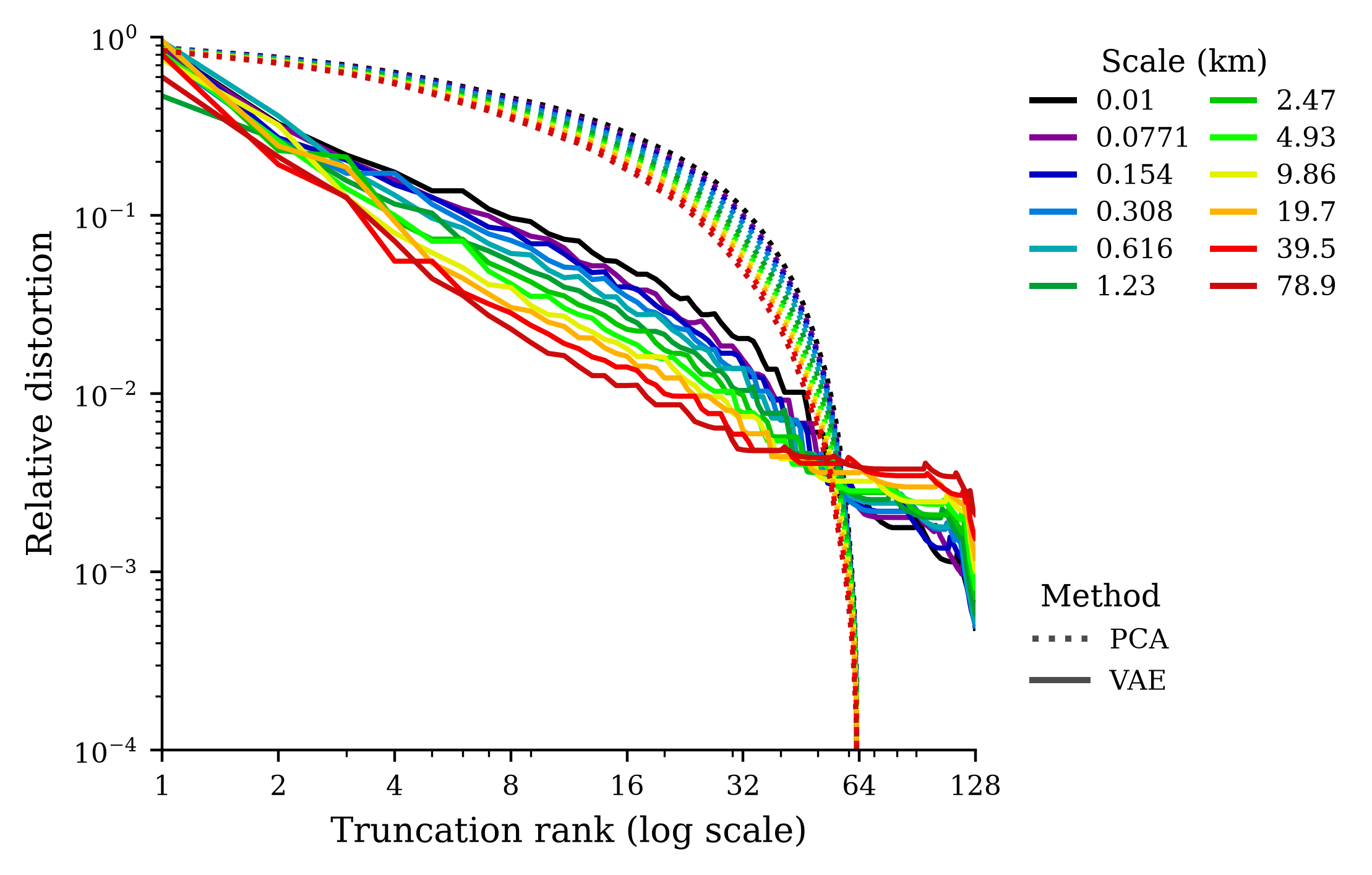}

}

\caption{\label{fig-appendix-rank-support}Held-out rank--distortion
curves for ball targets using the common VAE with eight 256-unit hidden
layers in each of the encoder and decoder and 128 latent coordinates
(solid), compared with PCA (dotted). Color denotes scale. Ranks \(K>64\)
show the overcomplete VAE tail.}

\end{figure}

The complete rank--distortion envelopes in
Figure~\ref{fig-appendix-rank-support} complement the scale--tolerance
contours in Figure~\ref{fig-dimension-support}. Each VAE point freezes
the checkpoint and latent order selected on validation for that rank,
then reports held-out test relative distortion. PCA uses its
train-fitted basis under the same validation-selection/test-evaluation
protocol.

The reference is the mean of all 14,881 transformed training targets. On
an evaluation set \(S\), its distortion is
\(D_0(S)=|S|^{-1}\sum_{i\in S}\lVert x_i-\bar{x}_{\mathrm{train}}\rVert^2\).
The same predictor is used for VAE and PCA, with its distortion measured
on the observations in each curve. Both methods use all 14,881 training,
3,154 validation, and 3,264 test anchors. PCA's mean and ordered basis
are fitted on the full training set and frozen before validation or test
evaluation.

\FloatBarrier

\clearpage

\subsection{Tolerance slices across
support}\label{tolerance-slices-across-support}

\begin{figure}[htbp]

\centering{

\includegraphics[width=0.94\linewidth,height=\textheight,keepaspectratio]{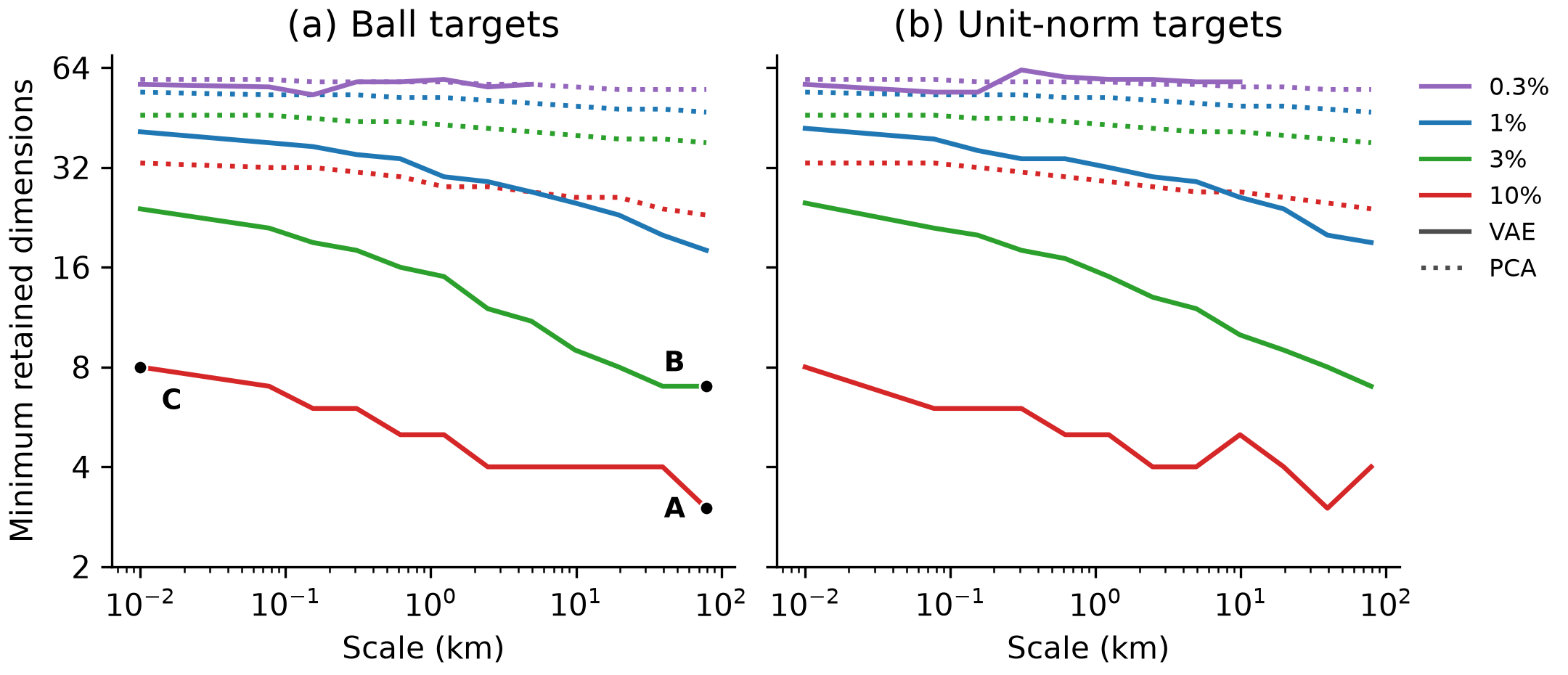}

}

\caption{\label{fig-appendix-tolerance-slices}Minimum
validation-selected dimension across scale for (a) ball and (b)
unit-norm targets. Solid lines show the VAE, dotted lines show PCA, and
color denotes tolerance. Dots mark model A, model B, and model C, the
ball-target descriptions used in subsequent figures.}

\end{figure}

The fixed-tolerance slices in Figure~\ref{fig-appendix-tolerance-slices}
give exact validation-selected counts at the measured supports. We
select and mark a few models for further analysis: model A retains three
coordinates at 78.9 km and 10\% tolerance, model B retains seven at 78.9
km and 3\%, and model C retains eight at native 10 m support and 10\%.
Table \ref{tbl-selected-model-distortions} reports the relative
distortions used for validation selection and those of the same frozen
models on test. Meeting a validation tolerance does not guarantee
meeting it on test.

\begin{longtable}[]{@{}lrrrr@{}}
\caption{Relative distortions of the validation-selected ball-target
models, with no selection on
test.}\label{tbl-selected-model-distortions}\tabularnewline
\toprule\noalign{}
Model & Support & Retained \(K\) & Validation \(D/D_0\) & Test
\(D/D_0\) \\
\midrule\noalign{}
\endfirsthead
\toprule\noalign{}
Model & Support & Retained \(K\) & Validation \(D/D_0\) & Test
\(D/D_0\) \\
\midrule\noalign{}
\endhead
\bottomrule\noalign{}
\endlastfoot
A & 78.9 km & 3 & 9.81\% & 12.55\% \\
B & 78.9 km & 7 & 2.60\% & 2.75\% \\
C & 10 m & 8 & 9.26\% & 9.65\% \\
\end{longtable}

\FloatBarrier

\clearpage

\subsection{Coarse-scale coordinate
maps}\label{coarse-scale-coordinate-maps}

\begin{figure}[htbp]

\centering{

\includegraphics[width=1\linewidth,height=\textheight,keepaspectratio]{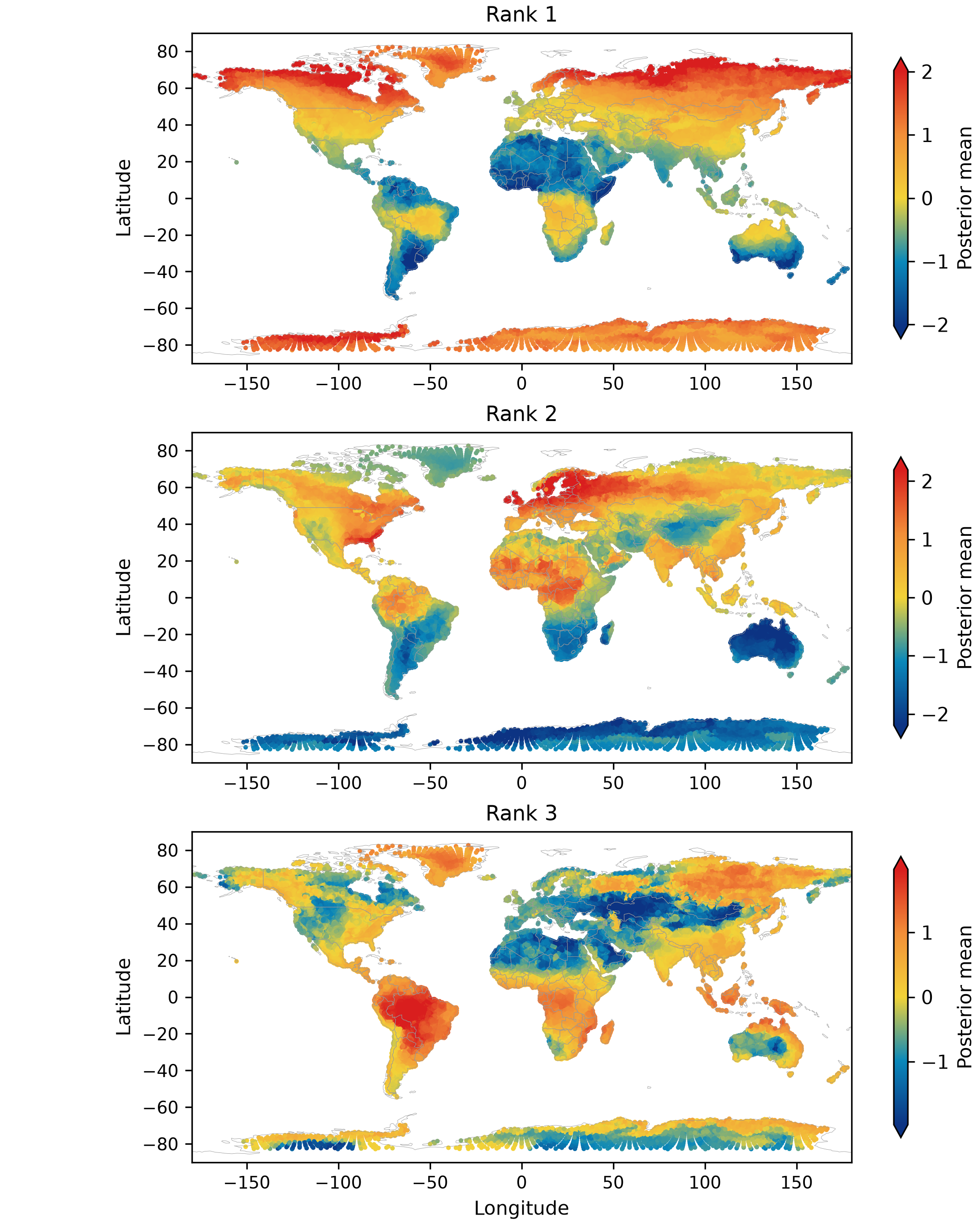}

}

\caption{\label{fig-appendix-k3-maps}Posterior means of the three
retained coordinates for model A (10\% validation tolerance at 78.9 km).
All 21,299 anchor centers are shown, in the original KL-rate order. Each
panel has its own symmetric color limits, clipped using the outer 2\%
tails. Latent signs are arbitrary. Blank areas have no retained
anchors.}

\end{figure}

The maps in Figure~\ref{fig-appendix-k3-maps} show the posterior means
of model A, selected at 78.9 km and 10\% tolerance
(Figure~\ref{fig-appendix-tolerance-slices}). These diagnostics use all
21,299 retained locations, pooling train, validation, and test.
Checkpoints, coordinate subsets, and their validation KL-rate order
remain frozen.

\FloatBarrier

\subsection{Environmental associations at coarse
support}\label{environmental-associations-at-coarse-support}

\begin{figure}[htbp]

\centering{

\includegraphics[width=1\linewidth,height=\textheight,keepaspectratio]{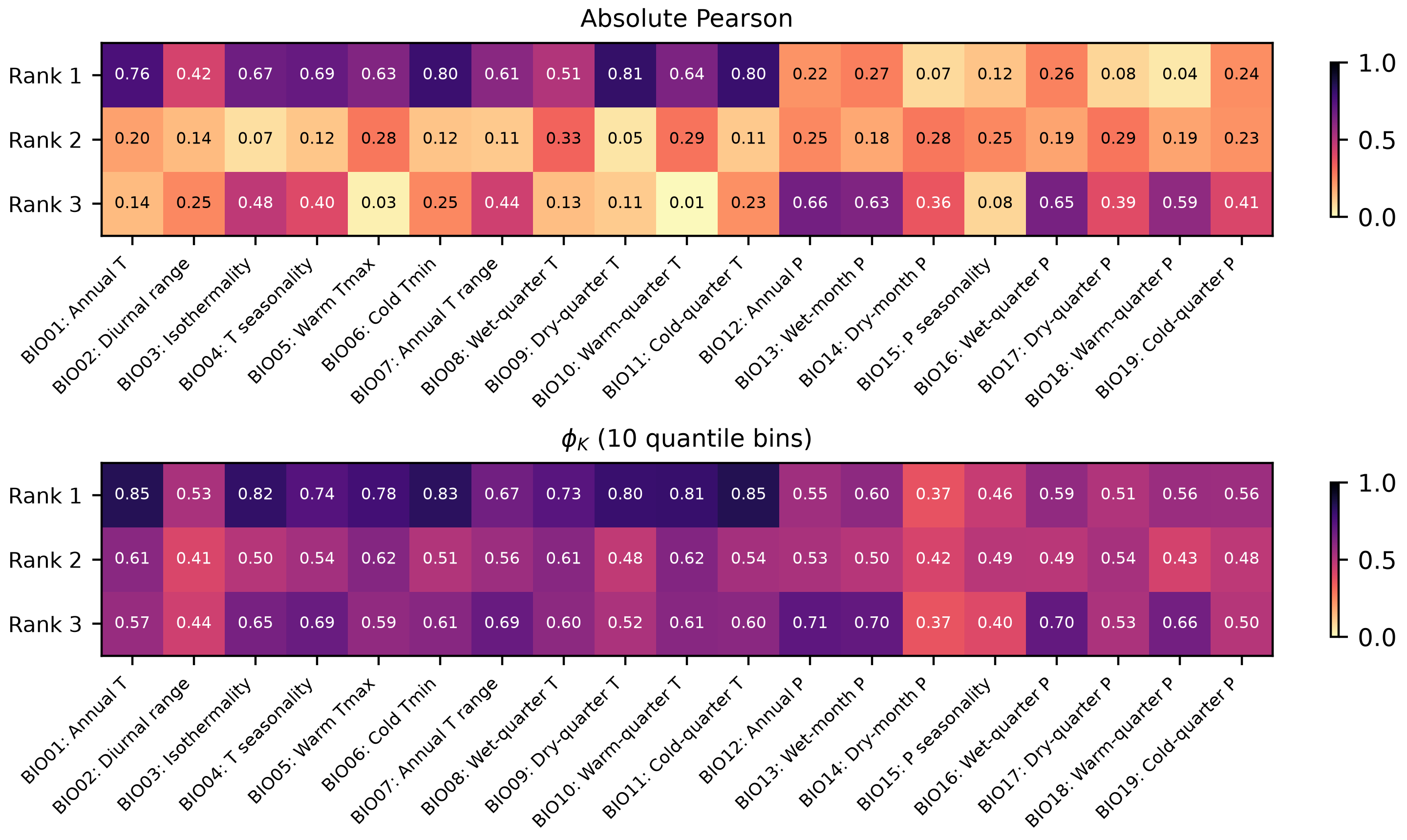}

}

\caption{\label{fig-appendix-k3-climate}Absolute Pearson and \(\phi_K\)
associations between the three retained posterior means of model A and
polygon-averaged WorldClim bioclimatic variables. Model A was selected
at 78.9 km and 10\% tolerance. T denotes temperature and P
precipitation. Ranks follow validation KL rate. Correlations use all
dataset splits. Rank 1 associates mainly with temperature, and rank 3
with annual and wet-season precipitation. Rank 2 has weak Pearson
correlations but appreciable \(\phi_K\) associations across both
groups.}

\end{figure}

\begin{figure}[htbp]

\centering{

\includegraphics[width=1\linewidth,height=\textheight,keepaspectratio]{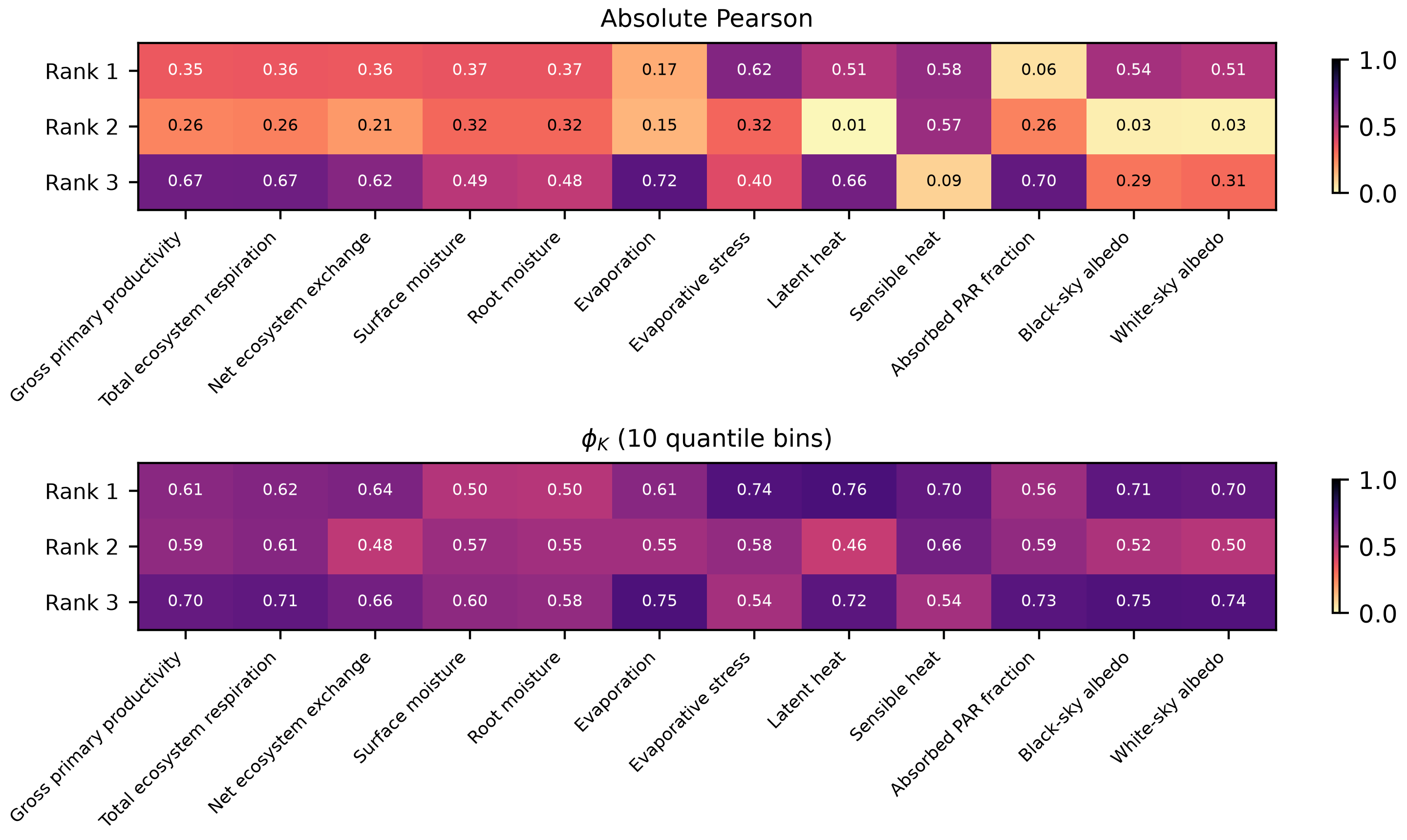}

}

\caption{\label{fig-appendix-k3-biosphere}Associations between the three
retained posterior means of model A and the twelve biosphere
climatologies. Missing-data filtering is applied independently for each
probe. Rank 1 associates with evaporative stress and heat fluxes, while
rank 3 associates with water and carbon-cycle quantities. The probes
overlap environmentally and do not identify separate physical
variables.}

\end{figure}

Figures \ref{fig-appendix-k3-climate}--\ref{fig-appendix-k3-biosphere}
compare the three posterior means of model A with environmental probes,
pooling all dataset splits as in the maps of
Figure~\ref{fig-appendix-k3-maps}. WorldClim 2.1 BIO01--BIO19 describe
1970--2000 climate at 10 arcminutes \citep{fick2017worldclim2}. We
aggregate them over the saved polygons with approximate area weights.
Biosphere probes use duration-weighted 2001--2011 climatologies from the
0.25-degree, eight-day dataset of \citet{kraemer2020biosphere}, sampled
at the original 256 descendant locations. Each biosphere cell requires
at least 80\% temporal coverage, and each polygon requires at least 80\%
valid sampled locations. Depending on the variable, 16,484--19,752
locations remain. These climatologies probe persistent spatial patterns,
not contemporaneous conditions in the 2024 embeddings. The biosphere
labels GPP, TER, NEE, and fAPAR denote gross primary productivity, total
ecosystem respiration, net ecosystem exchange, and the fraction of
absorbed photosynthetically active radiation (PAR). Black-sky and
white-sky albedo describe direct and diffuse illumination, respectively.

We report absolute Pearson correlation and the bounded nonlinear
association coefficient \(\phi_K\) \citep{baak2020correlation}. The
latter uses a noise-corrected contingency-table statistic calibrated to
Gaussian correlation, with ten marginal quantile bins and duplicate
edges merged. It can detect nonlinear associations missed by Pearson
correlation.

\FloatBarrier

\FloatBarrier

\clearpage

\subsection{Coordinate associations across
tolerance}\label{coordinate-associations-across-tolerance}

\begin{figure}[htbp]

\centering{

\includegraphics[width=1\linewidth,height=\textheight,keepaspectratio]{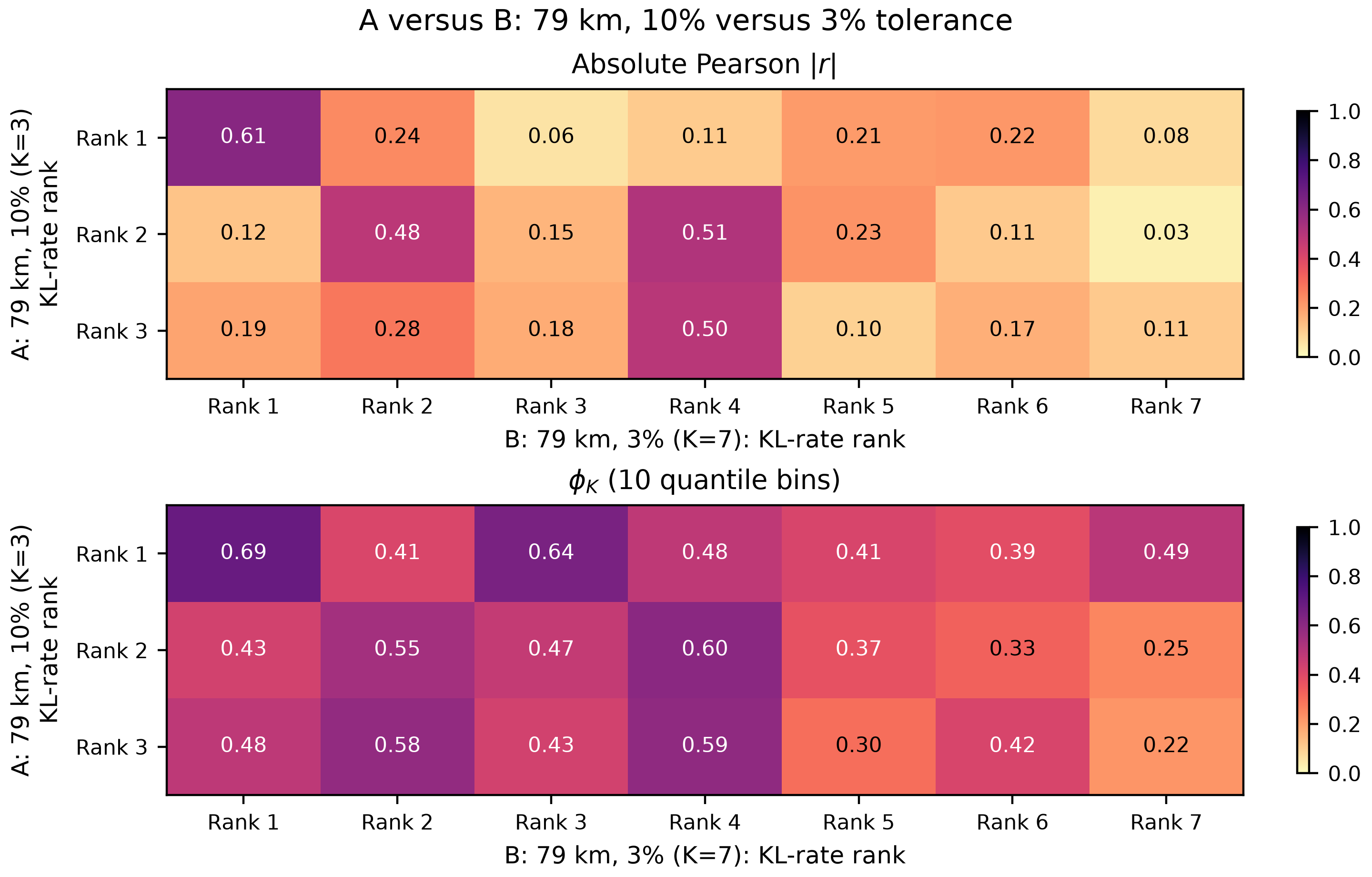}

}

\caption{\label{fig-appendix-cross-tolerance}Associations between model
A (three coordinates, 10\% tolerance) and model B (seven coordinates,
3\% tolerance), both at 78.9 km. Columns and rows follow each model's
own frozen KL-rate order. Both panels use all 21,299 matched locations
and a fixed zero-to-one color scale.}

\end{figure}

Figures
\ref{fig-appendix-cross-tolerance}--\ref{fig-appendix-cross-scale-seven-eight}
compare the posterior means of independently fitted ball-target models
at all 21,299 matched locations. Absolute correlations remove the
arbitrary sign of each coordinate. The \(\phi_K\) calculation uses the
same ten marginal quantile bins as the environmental probes.

At 78.9 km, model A (10\% tolerance, three coordinates) and model B
(3\%, seven coordinates) show a mixed correspondence
(Figure~\ref{fig-appendix-cross-tolerance}). Rank 1 of model A
associates with ranks 1 and 3 of model B. The A1--B3 pairing has
substantial \(\phi_K=0.64\) despite weak Pearson correlation
(\(|r|=0.06\)), illustrating the usefulness of the nonlinear measure.
Ranks 2 and 3 of model A both associate with ranks 2 and 4 of model B.
The compact description thus tracks variation distributed across several
coordinates of the tighter model.

\FloatBarrier

\clearpage

\subsection{Coordinate associations across spatial
support}\label{coordinate-associations-across-spatial-support}

\begin{figure}[htbp]

\centering{

\includegraphics[width=1\linewidth,height=\textheight,keepaspectratio]{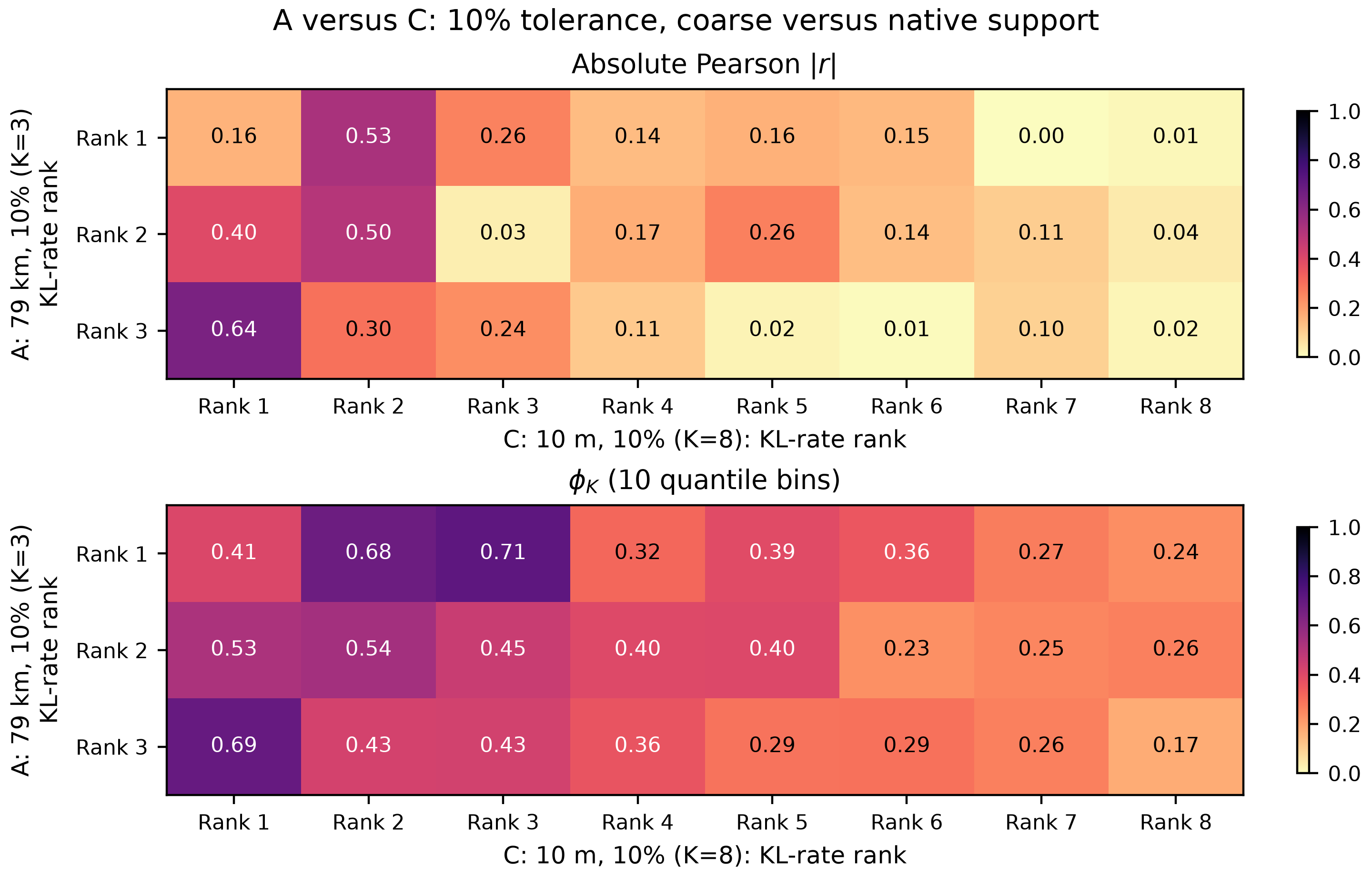}

}

\caption{\label{fig-appendix-cross-scale}Associations between model A
(three retained coordinates at 78.9 km) and model C (eight at native
resolution), both selected at 10\% tolerance. Each native embedding is
sampled at the corresponding polygon's anchor center. All 21,299 matched
locations enter both panels. Each model retains its own frozen KL-rate
order, and color limits match Figure
\ref{fig-appendix-cross-tolerance}.}

\end{figure}

At fixed 10\% tolerance, Figure~\ref{fig-appendix-cross-scale} pairs
model A at 78.9 km with model C at native resolution. In the \(\phi_K\)
panel, the strongest associations are concentrated in the first three
native coordinates, suggesting that the compact coarse description
remains related to the leading native variables. Their order changes:
coarse rank 3 associates most strongly with native rank 1, while coarse
rank 1 has its strongest Pearson association with native rank 2 and its
strongest \(\phi_K\) association with native rank 3. The native
representation encodes the single 10 m embedding at each anchor center,
while the coarse representation encodes the 256-sample polygon mean.
Thus the comparison concerns geographic associations between
descriptions with different spatial supports, not the average of native
latent coordinates over each polygon.

\FloatBarrier

\clearpage

\begin{figure}[htbp]

\centering{

\includegraphics[width=1\linewidth,height=\textheight,keepaspectratio]{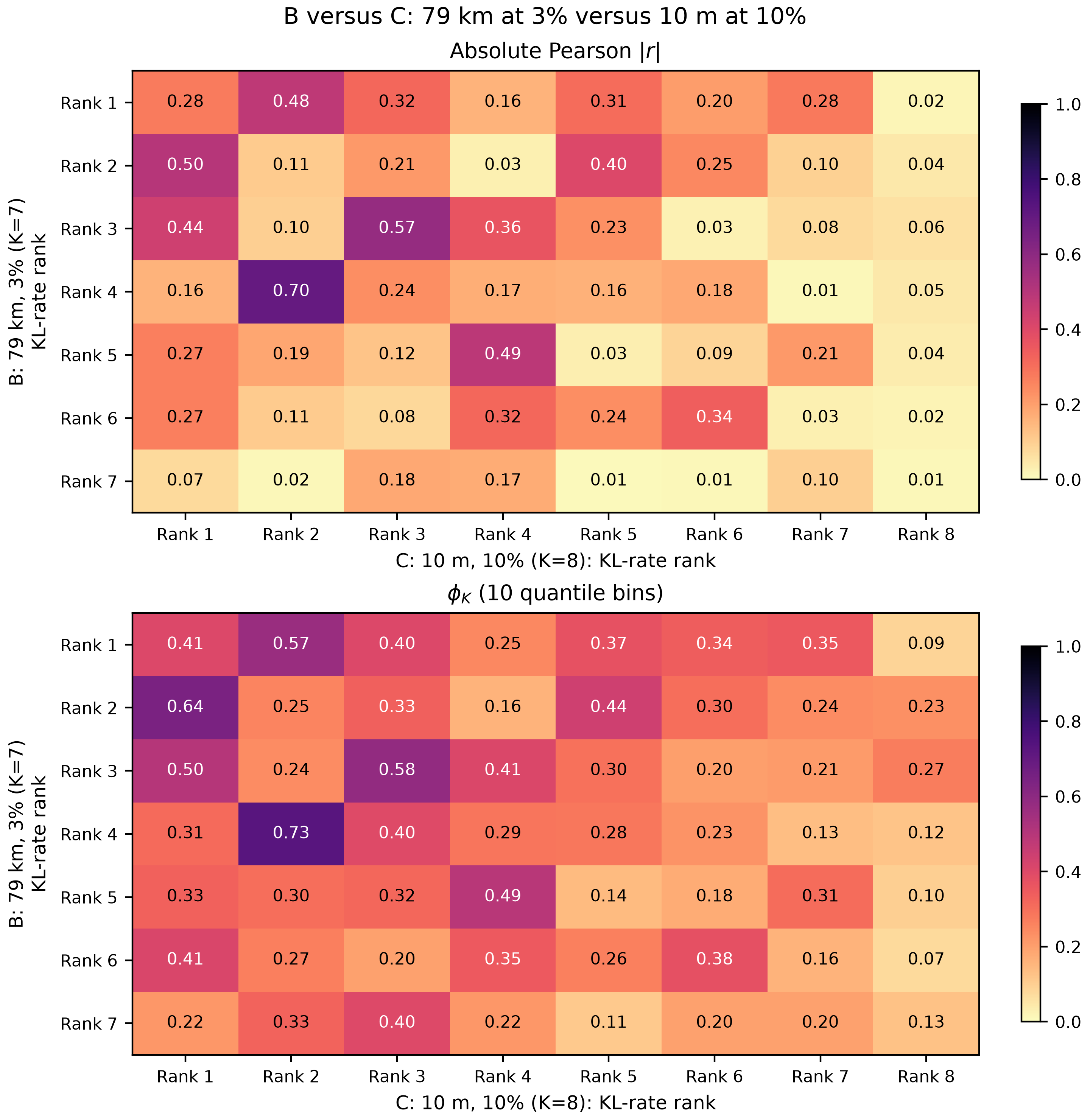}

}

\caption{\label{fig-appendix-cross-scale-seven-eight}Associations
between model B (seven retained coordinates at 78.9 km and 3\%
tolerance) and model C (eight at native resolution and 10\% tolerance).
All 21,299 matched locations enter both panels. The pairing uses the
native embedding at each anchor center and the corresponding
polygon-mean embedding, as in Figure \ref{fig-appendix-cross-scale}.
Coordinates retain each model's frozen KL-rate order, and color limits
match the other comparisons.}

\end{figure}

The comparison of model B and model C in
Figure~\ref{fig-appendix-cross-scale-seven-eight} pairs descriptions
with similar coordinate counts: seven at 78.9 km and eight at native
resolution. It changes both tolerance and spatial support, so it does
not isolate the effect of scale. Several leading coordinates show a
structured correspondence in both Pearson and \(\phi_K\). Ranks 1 and 2
of model B associate most strongly with ranks 2 and 1 of model C,
respectively, suggesting a partial reversal of their KL-rate order. Rank
3 of model B associates most strongly with rank 3 of model C, while its
rank 5 associates most strongly with rank 4. Rank 2 of model C
associates with both ranks 1 and 4 of model B, so the correspondence is
not simply a permutation. These patterns suggest shared structure across
supports, despite reordering and redistribution among coordinates.

\FloatBarrier

\clearpage

\subsection{Approximate collapse and effective
dimension}\label{approximate-collapse-and-effective-dimension}

Let \(e(K,\ell)\) denote the lowest relative validation distortion
attained with at most \(K\) coordinates, using the frozen KL orders and
per-rank checkpoint choices. At tolerance \(\varepsilon\), the effective
dimension is

\begin{equation}\protect\phantomsection\label{eq-effective-dimension-definition}{
d_{\mathrm{eff}}(\varepsilon,\ell)
=\min\{K:e(K,\ell)\leq\varepsilon\}.
}\end{equation}

\begin{figure}[htbp]

\centering{

\includegraphics[width=1\linewidth,height=\textheight,keepaspectratio]{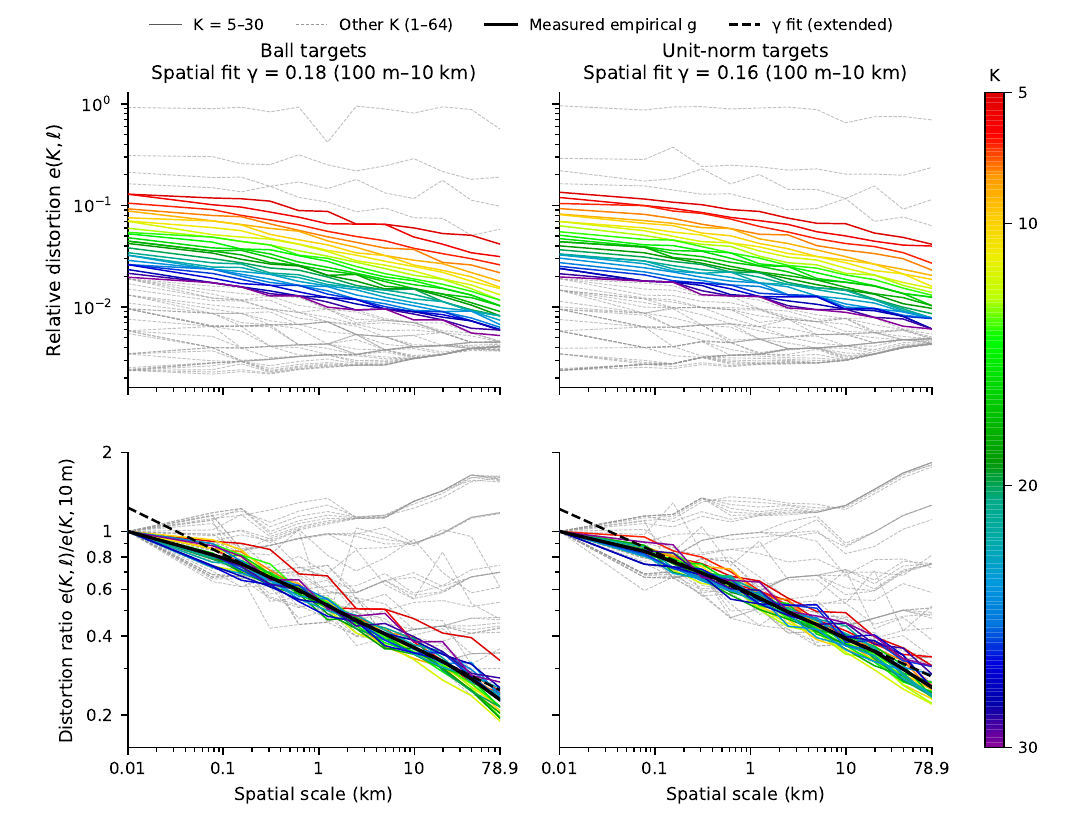}

}

\caption{\label{fig-appendix-curve-collapse}Validation scale curves for
\(K=1\)--64 before (top) and after (bottom) division by the measured
native 10 m relative distortion at each \(K\). Solid colored curves span
the empirical-profile interval \(K=5\)--30, and other ranks are gray
dashed. The bottom row shows the geometric-mean profile at all twelve
measured supports and its separate free-amplitude spatial power law,
fitted over 100 m--10 km and extended across the displayed scale range.
Portions outside the fitting interval are extrapolations.}

\end{figure}

We ask whether the rank--distortion curves at different supports share a
common dependence on \(K\), apart from a scale-dependent factor. Take
native support \(\ell_\star=10\,\mathrm{m}\) as the reference. The top
row of Figure~\ref{fig-appendix-curve-collapse} shows the unrescaled
scale curves for \(K=1\)--64. The bottom row divides each curve by its
measured native relative distortion \(e(K,\ell_\star)\). For the
intermediate range \(K=5\)--30, define the empirical spatial profile

\begin{equation}\protect\phantomsection\label{eq-empirical-spatial-profile}{
g(\ell)
=\exp\!\left[
\frac{1}{26}\sum_{K=5}^{30}
\log\frac{e(K,\ell)}{e(K,\ell_\star)}
\right].
}\end{equation}

All twelve measured supports enter this profile, with
\(g(\ell_\star)=1\) by definition. Ranks outside \(K=5\)--30 remain
visible as gray dashed curves.

In the intermediate rank range, the scale curves follow similar spatial
dependence and their native-normalized ratios lie close together. This
suggests that averaging mainly rescales the amplitude of a common
rank--distortion shape:

\begin{equation}\protect\phantomsection\label{eq-distortion-factorization}{
e(K,\ell)\approx e(K,\ell_\star)\,g(\ell).
}\end{equation}

The scale curves in Figure \ref{fig-appendix-curve-collapse} and rank
curves in Figure \ref{fig-appendix-rank-collapse} display two views of
this separation:

\begin{equation}\protect\phantomsection\label{eq-two-rescalings}{
\frac{e(K,\ell)}{e(K,\ell_\star)}\approx g(\ell),
\qquad
\frac{e(K,\ell)}{g(\ell)}\approx e(K,\ell_\star).
}\end{equation}

Estimating \(g\) calibrates one average logarithmic offset per support
over \(K=5\)--30.

The function \(F\) in Equation~\ref{eq-effective-dimension-scaling} is
the measured native effective-dimension curve,
\(F(\varepsilon)=d_{\mathrm{eff}}(\varepsilon,\ell_\star)\). When the
tolerance crossing lies in the shared \(K=5\)--30 range, the effective
dimension can therefore be read from this curve at a rescaled tolerance:

\begin{equation}\protect\phantomsection\label{eq-native-dimension-rescaling}{
d_{\mathrm{eff}}(\varepsilon,\ell)
\approx F\!\left(\frac{\varepsilon}{g(\ell)}\right)
=d_{\mathrm{eff}}\!\left(
\frac{\varepsilon}{g(\ell)},\ell_\star
\right).
}\end{equation}

At fixed \(K\), the measured coarse-support factors \(g(\ell)<1\) reduce
relative distortion within this range. At fixed tolerance,
\(\varepsilon/g(\ell)>\varepsilon\) means that fewer coordinates can
suffice. Coarse-graining therefore remains useful within the shared
middle interval. Low-rank departures indicate additional rank dependence
in its effect.

As a separate local summary, the empirical spatial profile is
approximately straight in log--log coordinates between 100 m and 10 km.
With reference scale \(\ell_0=1\,\mathrm{km}\), we fit

\begin{equation}\protect\phantomsection\label{eq-spatial-power-fit}{
\log g(\ell)\approx b-\gamma\log(\ell/\ell_0),
}\end{equation}

so \(g(\ell)\propto(\ell/\ell_0)^{-\gamma}\) within that window. The
intercept \(b\) is fitted using the seven measured supports between 100
m and 10 km. Figure \ref{fig-appendix-curve-collapse} shows this line at
its fitted amplitude across the displayed scale range, with the portions
outside 100 m--10 km extrapolated. Inside the fitting window, the solid
empirical profile and dashed power law nearly coincide. The rank
collapse in Figure \ref{fig-appendix-rank-collapse} uses the measured
\(g\) at all supports, without reference to this power-law
approximation.

\FloatBarrier

\subsubsection{Native rank dependence}\label{native-rank-dependence}

\begin{figure}[htbp]

\centering{

\includegraphics[width=1\linewidth,height=\textheight,keepaspectratio]{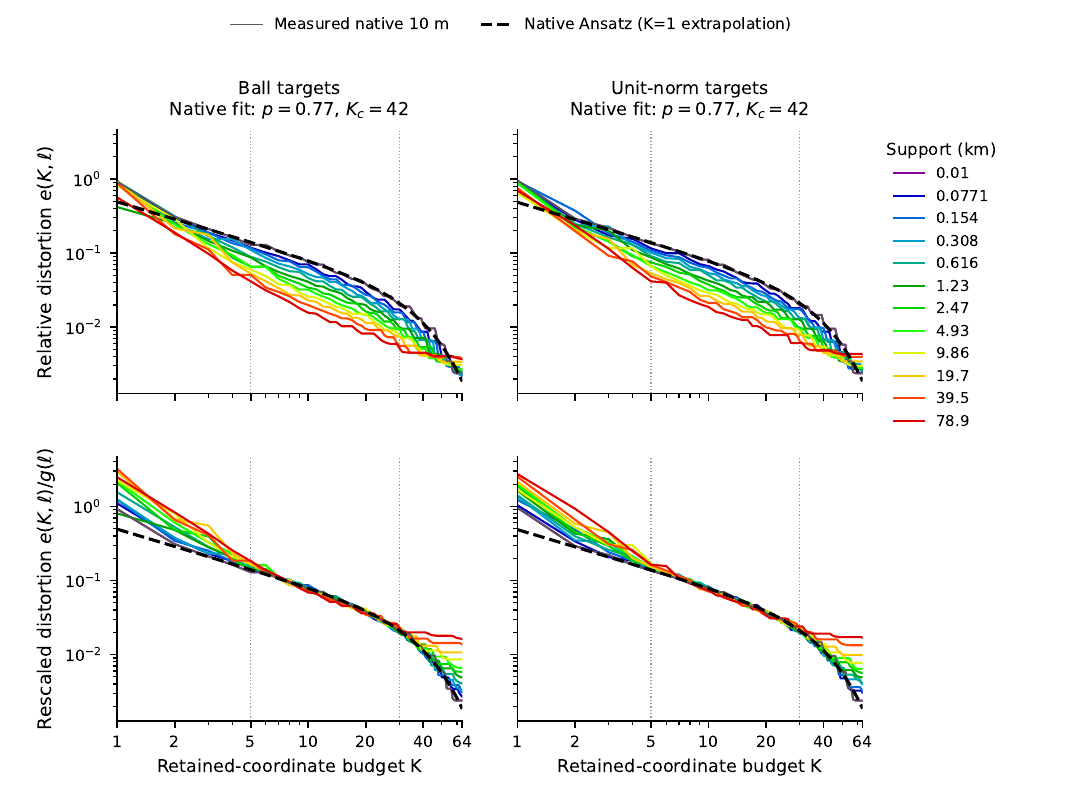}

}

\caption{\label{fig-appendix-rank-collapse}Validation rank--distortion
curves for \(K=1\)--64 before (top) and after (bottom) division by the
empirical \(K=5\)--30 spatial profile, with \(g(10\,\mathrm{m})=1\).
Color denotes support, from native violet to red at 78.9 km. Vertical
guides bound the profile-estimation interval. The native reference and
its black dashed native Ansatz are unchanged between rows because the
native spatial factor is exactly one. The fit uses native \(K=2\)--64
only, with the dashed extension to \(K=1\) shown as extrapolation.}

\end{figure}

The shape of the native rank curve can also be approximated separately.
Figure \ref{fig-appendix-rank-collapse} shows curvature in log--log
coordinates, with progressively steeper rank dependence before the
attained high-rank plateau. A smooth empirical Ansatz approximates the
measured native relative distortions:

\begin{equation}\protect\phantomsection\label{eq-native-rank-ansatz}{
\widehat e(K,\ell_\star)=A K^{-p}\exp[-(K/K_c)^2].
}\end{equation}

We fit it directly to the native validation relative distortions over
\(K=2\)--64, weighting each integer \(K\) equally in squared natural-log
residual. This fit uses native distortions alone, so \(g(\ell)\) or
\(\gamma\) do not enter the fit. The large change between \(K=1\) and
\(K=2\) precedes the smoother regime, so \(K=1\) remains visible but is
excluded from fitting. The dashed continuation to \(K=1\) is displayed
extrapolation only. The local logarithmic slope,

\begin{equation}\protect\phantomsection\label{eq-local-rank-slope}{
\frac{d\log\widehat e(K,\ell_\star)}{d\log K}=-p-2(K/K_c)^2,
}\end{equation}

becomes more negative as \(K\) increases. A straight power law can
therefore approximate an intermediate segment, with a signed exponent
that depends on the rank range. The fitted \(K_c\) controls the
curvature of this rank dependence and does not by itself identify a
critical or intrinsic dimension. The approximate scaling relation
neither aligns latent coordinates nor establishes criticality or a
renormalization-group flow.

\begin{longtable}[]{@{}lrrrr@{}}
\caption{Spatial exponent fitted to the empirical \(K=5\)--30 profile
over 100 m--10 km, followed by separate native-Ansatz parameters fitted
over \(K=2\)--64. \(K_c\) is rounded to two significant
digits.}\label{tbl-collapse-parameters}\tabularnewline
\toprule\noalign{}
Targets & Spatial \(\gamma\) & Native \(A\) & Native \(p\) & Native
\(K_c\) \\
\midrule\noalign{}
\endfirsthead
\toprule\noalign{}
Targets & Spatial \(\gamma\) & Native \(A\) & Native \(p\) & Native
\(K_c\) \\
\midrule\noalign{}
\endhead
\bottomrule\noalign{}
\endlastfoot
Ball & 0.18 & 0.49 & 0.77 & 42 \\
Unit-norm & 0.16 & 0.49 & 0.77 & 42 \\
\end{longtable}

\FloatBarrier

\clearpage

\subsubsection{Closed-form dimension
approximation}\label{closed-form-dimension-approximation}

\begin{figure}[htbp]

\centering{

\includegraphics[width=1\linewidth,height=\textheight,keepaspectratio]{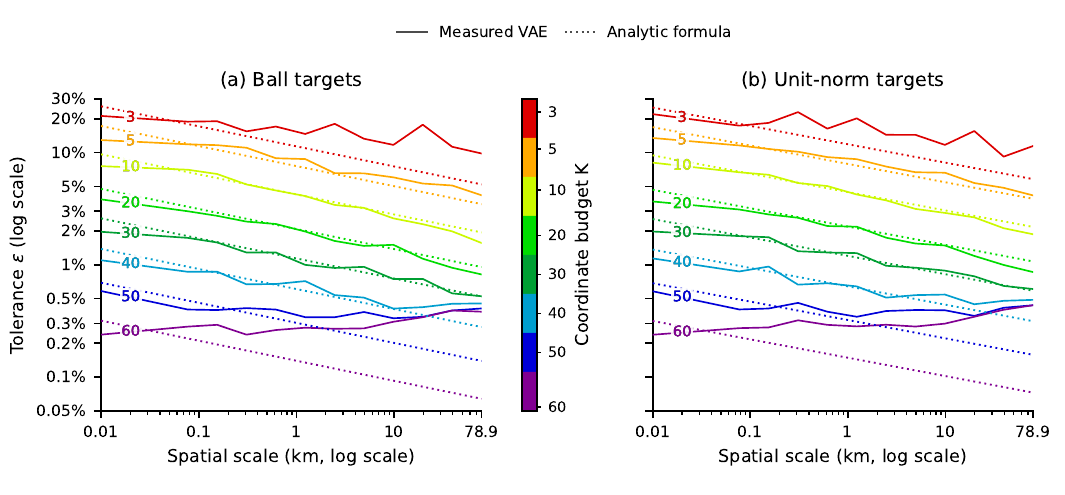}

}

\caption{\label{fig-appendix-analytic-dimension}Measured validation
scale--tolerance contours (solid) and the combined analytic
approximation (dotted), for ball targets (left) and unit-norm targets
(right). The \(K=3\) comparison exposes departures at low rank, while
\(K=50\) and \(60\) expose departures in the high-rank tail.}

\end{figure}

Figure \ref{fig-appendix-analytic-dimension} compares the measured
scale--tolerance contours with an analytic approximation. Combining the
native rank fit with the local spatial power law gives a fitted
approximation to the relative distortion attained with \(K\)
coordinates:

\begin{equation}\protect\phantomsection\label{eq-combined-distortion-approximation}{
e(K,\ell)\approx\widehat e(K,\ell_\star)\,
e^b\left(\frac{\ell}{\ell_0}\right)^{-\gamma}.
}\end{equation}

The native parameters are given in Table \ref{tbl-collapse-parameters}.
The fitted spatial amplitudes \(e^b\) at \(\ell_0=1\,\mathrm{km}\) are
0.54 for ball targets and 0.57 for unit-norm targets. The spatial fit
uses only 100 m--10 km data and is not constrained to equal the measured
\(g(\ell_\star)=1\) at native support. The native rank curve was fitted
separately.

We denote the continuous inverse of this fitted distortion curve, before
rounding to an integer coordinate count, by
\(\widetilde d_{\mathrm{eff}}\):

\begin{equation}\protect\phantomsection\label{eq-analytic-dimension-inverse}{
\widetilde d_{\mathrm{eff}}(\varepsilon,\ell)
=K_c\sqrt{\frac p2\,
W_0\!\left[
\frac{2}{pK_c^2}
\left(
\frac{A e^b}{\varepsilon}
\left(\frac{\ell}{\ell_0}\right)^{-\gamma}
\right)^{2/p}
\right]}.
}\end{equation}

Here \(W_0\) is the principal Lambert function, satisfying
\(W_0(x)e^{W_0(x)}=x\). Its argument is positive, and the fitted
distortion decreases with \(K\), so upward rounding gives the predicted
integer count. This fitted inverse is distinct from \(F\), which remains
the measured native effective-dimension curve.

Figure \ref{fig-appendix-analytic-dimension} displays the combined
approximation as dotted curves across the full spatial range, including
its extension outside the spatial fitting window. The \(K=3\) comparison
exposes departures from the shared shape at low rank. At \(K=50\) and
\(60\), measured distortion no longer decreases steadily with scale
while the formula continues to decrease, showing the limit of extending
the intermediate-rank spatial relation to the tail. The formula
summarizes the intermediate regime and its nearby crossings. It does not
replace the measured dimension counts or describe every rank and scale
equally well.

\begin{thebibliography}{15}
\providecommand{\natexlab}[1]{#1}
\providecommand{\url}[1]{\texttt{#1}}
\expandafter\ifx\csname urlstyle\endcsname\relax
  \providecommand{\doi}[1]{doi: #1}\else
  \providecommand{\doi}{doi: \begingroup \urlstyle{rm}\Url}\fi

\bibitem[Brown et~al.(2025)Brown, Kazmierski, Pasquarella, Rucklidge, Samsikova, Zhang, Shelhamer, Lahera, Wiles, Ilyushchenko, Gorelick, Zhang, Alj, Schechter, Askay, Guinan, Moore, Boukouvalas, and Kohli]{brown2025alphaearth}
Christopher~F. Brown, Michal~R. Kazmierski, Valerie~J. Pasquarella, William~J. Rucklidge, Masha Samsikova, Chenhui Zhang, Evan Shelhamer, Estefania Lahera, Olivia Wiles, Simon Ilyushchenko, Noel Gorelick, Lihui~Lydia Zhang, Sophia Alj, Emily Schechter, Sean Askay, Oliver Guinan, Rebecca Moore, Alexis Boukouvalas, and Pushmeet Kohli.
\newblock Alphaearth foundations: An embedding field model for accurate and efficient global mapping from sparse label data.
\newblock arXiv preprint, 2025.
\newblock URL \url{https://arxiv.org/abs/2507.22291}.
\newblock arXiv:2507.22291.

\bibitem[Weinberg(1979)]{weinberg1979phenomenological}
Steven Weinberg.
\newblock Phenomenological lagrangians.
\newblock \emph{Physica A}, 96\penalty0 (1--2):\penalty0 327--340, 1979.
\newblock \doi{10.1016/0378-4371(79)90223-1}.

\bibitem[Kingma and Welling(2014)]{kingma2014autoencoding}
Diederik~P. Kingma and Max Welling.
\newblock Auto-encoding variational bayes.
\newblock In \emph{2nd International Conference on Learning Representations (ICLR)}, 2014.
\newblock URL \url{https://arxiv.org/abs/1312.6114}.

\bibitem[Higgins et~al.(2017)Higgins, Matthey, Pal, Burgess, Glorot, Botvinick, Mohamed, and Lerchner]{higgins2017betavae}
Irina Higgins, Lo{\"i}c Matthey, Arka Pal, Christopher~P. Burgess, Xavier Glorot, Matthew~M. Botvinick, Shakir Mohamed, and Alexander Lerchner.
\newblock beta-vae: Learning basic visual concepts with a constrained variational framework.
\newblock In \emph{5th International Conference on Learning Representations (ICLR)}. OpenReview.net, 2017.
\newblock URL \url{https://openreview.net/forum?id=Sy2fzU9gl}.

\bibitem[Hirn(2026{\natexlab{a}})]{hirn2026posteriorcollapse}
Johannes Hirn.
\newblock Latent spectroscopy: Posterior collapse as a feature.
\newblock arXiv preprint, 2026{\natexlab{a}}.
\newblock URL \url{https://arxiv.org/abs/2605.22691v2}.
\newblock arXiv:2605.22691v2.

\bibitem[Hirn(2026{\natexlab{b}})]{hirn2026generative}
Johannes Hirn.
\newblock \texorpdfstring{$\beta$}{beta}-vaes as effective theories: Tolerance-dependent dimension.
\newblock arXiv preprint, 2026{\natexlab{b}}.
\newblock URL \url{https://arxiv.org/abs/2608.10599}.
\newblock arXiv:2608.10599.

\bibitem[Wilson(1971)]{wilson1971rg1}
Kenneth~G. Wilson.
\newblock Renormalization group and critical phenomena. i. renormalization group and the kadanoff scaling picture.
\newblock \emph{Physical Review B}, 4\penalty0 (9):\penalty0 3174--3183, 1971.
\newblock \doi{10.1103/PhysRevB.4.3174}.
\newblock URL \url{https://link.aps.org/doi/10.1103/PhysRevB.4.3174}.

\bibitem[Rahman et~al.(2026)Rahman, Barrett, and Last]{rahman2026characterizing}
Mashrekur Rahman, Samuel~J. Barrett, and Christina Last.
\newblock Characterizing {AlphaEarth} embedding geometry for agentic environmental reasoning.
\newblock arXiv preprint, 2026.
\newblock URL \url{https://arxiv.org/abs/2604.18715}.
\newblock arXiv:2604.18715.

\bibitem[Rao et~al.(2026)Rao, Ru{\ss}wurm, Klemmer, and Rolf]{rao2026intrinsicdimensionearth}
Arjun Rao, Marc Ru{\ss}wurm, Konstantin Klemmer, and Esther Rolf.
\newblock Measuring the intrinsic dimension of earth representations.
\newblock In \emph{International Conference on Learning Representations}, 2026.
\newblock \doi{10.48550/arXiv.2511.02101}.
\newblock URL \url{https://arxiv.org/abs/2511.02101}.
\newblock arXiv:2511.02101.

\bibitem[Rahman(2026)]{rahman2026alphaearthinterpretable}
Mashrekur Rahman.
\newblock Physically interpretable {AlphaEarth} foundation model embeddings enable {LLM}-based land surface intelligence.
\newblock \emph{Remote Sensing Applications: Society and Environment}, 42:\penalty0 102045, 2026.
\newblock \doi{10.1016/j.rsase.2026.102045}.
\newblock URL \url{https://doi.org/10.1016/j.rsase.2026.102045}.

\bibitem[Benavides-Martinez et~al.(2026)Benavides-Martinez, Guthrie, Arias, Garces-Gomez, Guzman-Alvis, Portilla-Cabrera, Mondal, Allyn, and Ganguly]{benavides2026whatonearthisalphaearth}
Ivan~Felipe Benavides-Martinez, Justin Guthrie, Jhon~Edwin Arias, Yeison~Alberto Garces-Gomez, Angela~Ines Guzman-Alvis, Cristiam~Victoriano Portilla-Cabrera, Somnath Mondal, Andrew~J. Allyn, and Auroop~R. Ganguly.
\newblock What on earth is {AlphaEarth}? hierarchical structure and functional interpretability for global land cover.
\newblock arXiv preprint, 2026.
\newblock URL \url{https://arxiv.org/abs/2603.16911}.
\newblock arXiv:2603.16911.

\bibitem[Barenboim et~al.(2024)Barenboim, Del~Debbio, Hirn, and Sanz]{barenboim2024music}
Gabriela Barenboim, Luigi Del~Debbio, Johannes Hirn, and Ver{\'o}nica Sanz.
\newblock Exploring how a generative {AI} interprets music.
\newblock \emph{Neural Computing and Applications}, 36\penalty0 (27):\penalty0 17007--17022, 2024.
\newblock \doi{10.1007/s00521-024-09956-9}.
\newblock URL \url{https://doi.org/10.1007/s00521-024-09956-9}.

\bibitem[Fick and Hijmans(2017)]{fick2017worldclim2}
Stephen~E. Fick and Robert~J. Hijmans.
\newblock Worldclim 2: new 1-km spatial resolution climate surfaces for global land areas.
\newblock \emph{International Journal of Climatology}, 37\penalty0 (12):\penalty0 4302--4315, 2017.
\newblock \doi{10.1002/joc.5086}.
\newblock URL \url{https://doi.org/10.1002/joc.5086}.

\bibitem[Kraemer et~al.(2020)Kraemer, Camps-Valls, Reichstein, and Mahecha]{kraemer2020biosphere}
Guido Kraemer, Gustau Camps-Valls, Markus Reichstein, and Miguel~D. Mahecha.
\newblock Summarizing the state of the terrestrial biosphere in few dimensions.
\newblock \emph{Biogeosciences}, 17:\penalty0 2397--2424, 2020.
\newblock \doi{10.5194/bg-17-2397-2020}.

\bibitem[Baak et~al.(2020)Baak, Koopman, Snoek, and Klous]{baak2020correlation}
M.~Baak, R.~Koopman, H.~Snoek, and S.~Klous.
\newblock A new correlation coefficient between categorical, ordinal and interval variables with pearson characteristics.
\newblock \emph{Computational Statistics \& Data Analysis}, 152:\penalty0 107043, 2020.
\newblock \doi{10.1016/j.csda.2020.107043}.

\end{thebibliography}
\end{document}